\documentclass[10pt,twocolumn,letterpaper]{article}

\usepackage[pagenumbers]{cvpr}      %

\usepackage{url}
\usepackage{nicefrac}
\usepackage[dvipsnames]{xcolor}
\usepackage{wrapfig}
\usepackage{enumitem}
\usepackage{amsmath,bm}
\usepackage{dsfont}
\usepackage{amssymb}
\usepackage{amsfonts}
\usepackage{dcolumn}
\usepackage{tabu}
\usepackage{array}
\usepackage{xspace}
\usepackage{makecell}
\usepackage{colortbl}
\usepackage{multirow}
\usepackage{soul}
\usepackage{changepage,threeparttable}
\usepackage{algorithm}
\usepackage{algpseudocode}
\usepackage[vlined,linesnumbered,ruled,algo2e]{algorithm2e}
\usepackage{mathtools}
\usepackage{amsthm}

\usepackage{cuted}
\usepackage{fontawesome5}
\theoremstyle{plain}

\theoremstyle{definition}

\theoremstyle{remark}

\newcommand{\cmark}{\textcolor{ForestGreen}{\(\checkmark\)}}
\newcommand{\xmark}{\textcolor{red}{\(\times\)}}

\definecolor{myorange}{RGB}{230,120,20}

\newcommand{\bestmark}{%
    \colorbox{bestbg}{\scriptsize Best}}
\newcommand{\secondmark}{%
    \colorbox{secondbg}{\scriptsize Second-best}}
\definecolor{cvprblue}{rgb}{0.21,0.49,0.74}
\usepackage[pagebackref,breaklinks,colorlinks,allcolors=cvprblue]{hyperref}
\usepackage[capitalize,noabbrev]{cleveref}

\title{Fluid-Gen-Zero: Grounding Pretrained Video Generators in Physics without Training}

\author{
Hong Huang$^{1}$ \enspace
Yuqiu Liu$^{1}$ \enspace
Chenyu You$^{2}$ \enspace
Daniel Martin$^{3}$ \enspace
Chuhang Zou$^{4\dagger}$ \enspace
Wuyang Chen$^{1\dagger}$
\\
$^{1}$Simon Fraser University \quad
$^{2}$Stony Brook University \quad
 \\
$^{3}$Lawrence Berkeley National Laboratory \quad
$^{4}$Meta Reality Labs
\\
$^{\dagger}$Corresponding authors
\\[0.3em]
\href{https://hongh0.github.io/Fluid-Gen-Zero/}
{\faGlobe~Project Page}
}

\begin{document}
\twocolumn[{%
\renewcommand\twocolumn[1][]{#1}%
\maketitle
\begin{center}
    \centering
    \vspace{-1.5em}
    \includegraphics[width=\textwidth]{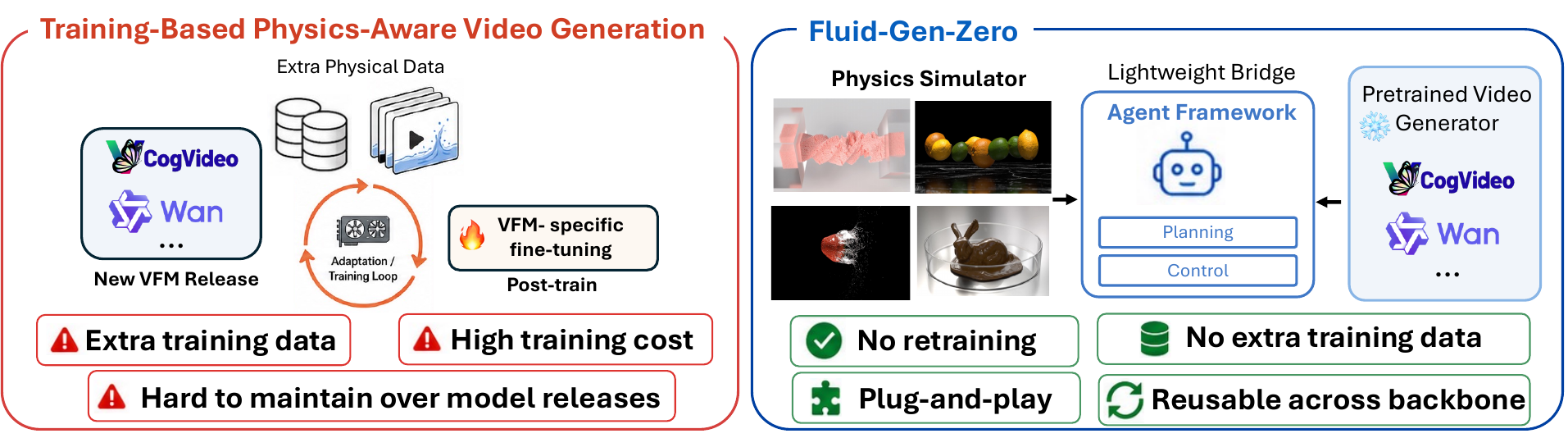}
    \vspace{-1.5em}
    \captionof{figure}{\textbf{Training-based \emph{vs.}\ Training-free physics-aware video generation.}
    \emph{(Left)} Existing approaches rely on architecture-coupled post-training to inject physical priors, creating a rigid dependency on specific model backbones.
    \emph{(Right)} \textbf{Fluid-Gen-Zero} decouples physical reasoning from appearance synthesis. By introducing physics awareness entirely at inference time, it offers a plug-and-play solution compatible with rapidly evolving video foundation models.}
    \label{fig:motivation}
\end{center}
\vspace{0.5em}
}]
\begin{abstract}
We present \textbf{Fluid-Gen-Zero}, a training-free framework for physics-aware fluid-object interaction video generation that decouples physical reasoning from appearance synthesis. Our key insight is to delegate motion dynamics to a physics simulator while preserving the appearance modeling capacity of pretrained video generators. We bridge these two domains through a two-level agentic workflow: generation-time planning, where a vision-language model (VLM) agent interprets intent and the simulation rollout to organize generation clips, and latent-space guidance, which injects simulation signals into denoising through region-aware latent wrapping. This plug-and-play design is compatible with current video foundation models. We further introduce a benchmark for fluid-object interaction video generation. Across Tora (CogVideoX-based), VACE and WanMove (Wan-based), Fluid-Gen-Zero consistently improves simulation alignment, reducing object trajectory error by \textbf{26.7\%--81.5\%} and fluid fEPE (fluid flow endpoint error) by \textbf{67.9\%--84.0\%}, while largely preserving perceptual quality. In a human preference study, raters favor Fluid-Gen-Zero in \textbf{55.1\%--74.4\%} of same-backbone comparisons across three backbones, and in \textbf{90.4\%--94.2\%} of comparisons against simulation-based methods. Code and data will be released upon acceptance.

\end{abstract}

\section{Introduction}

Video foundation models (VFMs) achieve remarkable visual fidelity~\cite{cogvideox,agarwal2025cosmos,sora,hunyuanvideo2025,veo3,kling}, yet they are not inherently grounded in physical laws. This challenge is particularly evident in fluid-object interactions, where complex, deformable dynamics cannot be reliably inferred from text or standard diffusion priors~\cite{stam2023stable,bridson2015fluid,hu2018moving}. Consequently, state-of-the-art generators frequently produce visually appealing but physically implausible motion~\cite{meng2024towards,bansal2024videophy,wang2025wisa}.

To bridge this gap, recent works inject physical priors through fine-tuning~\cite{meng2024towards,bansal2024videophy,wang2025wisa,li2025pisa}, preference optimization~\cite{physhpo,cai2025phygdpo,xue2025phyt2v}, or auxiliary objectives~\cite{li2025wonderplay,foo2026psivg,liu2026realwonder,zhan2026perpetualwonder,chen2025physgen3d,xie2025physanimator}.
While these methods improve physical grounding, their reliance on \textbf{architecture-coupled post-training} presents \textbf{scalability challenges}.
\emph{First}, updating the weights of massive video generators incurs prohibitive costs in computation and training data collection.
\emph{Second}, as VFMs scale and evolve at an unprecedented pace, maintaining architecture-coupled post-training paradigms becomes increasingly unsustainable.
An architecture-specific physics adapter trained today risks immediate obsolescence when a new VFM is released tomorrow, a dynamic reminiscent of the ``bitter lesson''~\cite{sutton2019bitter}.

This highlights the critical necessity of a \textbf{training-free} approach, as shown in \cref{fig:motivation}. By treating physics awareness as an inference-time capability, a training-free framework avoids costly retraining, preserves the base model's visual fidelity, and offers plug-and-play compatibility across future architectures. A sustainable solution must therefore \emph{decouple physical reasoning from the generator's weights}, prompting a fundamental question:

\begin{center}
\fbox{
    \parbox{0.86\linewidth}{
        \textit{How to bridge pretrained video generators and physics awareness without training?}
    }
}
\end{center}

In this work, we propose \textbf{Fluid-Gen-Zero}, a training-free framework that decouples physical reasoning from appearance synthesis, enabling plug-and-play physics awareness across production-level video generators in fluid scenes. Our intervention is logically based on two universally accessible handles: the generation timeline and the internal denoising trajectory.

Fluid-Gen-Zero leverages these handles at two complementary levels. \textbf{Generation-time planning} translates the structured intent record $\mathcal{I}$ and simulation rollout into a clip plan with clip-level prompts and protected event intervals. \textbf{Latent-space guidance} injects simulation signals into denoising through region-aware latent wrapping.
Together, these mechanisms recast physics awareness from a rigid training objective into a portable, inference-time capability. We summarize our contributions below:

\begin{itemize}[leftmargin=*, nosep]
    \item We propose \textbf{Fluid-Gen-Zero}, a training-free framework that introduces physics awareness through generation-time planning and latent-space guidance, providing plug-and-play compatibility with multiple pretrained video generators.
    \item Across Tora (CogVideoX-based), VACE, and WanMove (Wan-based) backbones, Fluid-Gen-Zero consistently improves simulation alignment, reducing object trajectory error by \textbf{26.7\%--81.5\%} and fluid fEPE by \textbf{67.9\%--84.0\%}, while largely preserving perceptual quality.
    \item We contribute a benchmark of 43 curated \textbf{fluid-object interaction} samples with 401-frame simulation rollouts, 3D particle trajectories, object and fluid masks, and projected simulation optical flow across 10 primary fluid categories.
\end{itemize}

\begin{figure*}[t]
    \centering
    \vspace{-0.5em}
    \includegraphics[width=\textwidth]{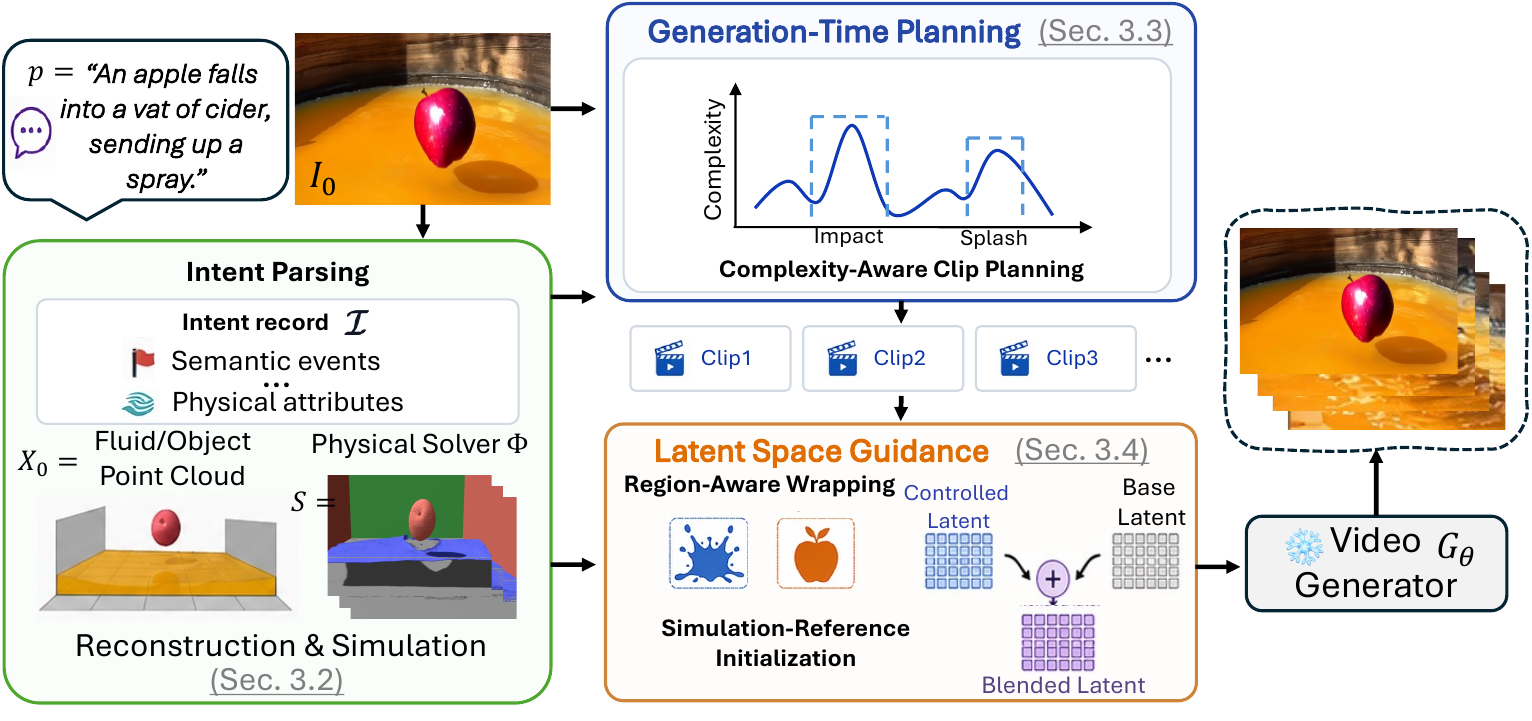}
    \vspace{-2.5em}
    \caption{\textbf{Fluid-Gen-Zero pipeline.} Given an input image and input prompt, we first parse the input, reconstruct a simulation-ready 3D scene, and roll out the corresponding fluid-object dynamics.
\emph{Generation-time planning} converts the simulation rollout into a clip plan with protected event intervals and clip-level prompts. \emph{Latent-space guidance} injects latent-control inputs into the denoising trajectory of a frozen video generator. }

\vspace{-1.5em}
\label{fig:pipeline}
\end{figure*}

\section{Related Work}

Physics-based simulators, from classical SPH/PBF and MPM solvers~\cite{monaghan1992smoothed,stomakhin2013material,bridson2015fluid,stam2023stable,hu2018moving} to engines such as MuJoCo, Taichi, Warp, and Genesis~\cite{todorov2012mujoco,hu2019taichi,Macklin_Warp_A_High-performance_2022,Genesis}, model fluid dynamics and multi-material coupling while preserving causality and contact, but they require explicit geometry and materials and do not synthesize open-vocabulary appearance. Video foundation models~\cite{cogvideox,hunyuanvideo2025,sora,veo3,kling,agarwal2025cosmos} provide high visual quality, and controllable variants condition on masks, depth, drags, or trajectories~\cite{wu2024draganything,li2025image,ling2024motionclone,namekata2024sgi2v}, as in MotionCtrl~\cite{wang2024motionctrl}, Tora~\cite{zhang2025tora}, MagicMotion~\cite{li2025magicmotion}, WanMove~\cite{chu2025wanmove}, and ATI~\cite{wang2025ati}, or use training-free schedules, as in MotionCraft~\cite{savantaira2024motioncraft} and Time-to-Move~\cite{singer2025time}; these image- or latent-space controls do not determine physically valid 3D dynamics. Physics-informed generators improve physical consistency through fine-tuning and preference optimization~\cite{meng2024towards,wang2025wisa,li2025pisa,physhpo,cai2025phygdpo}, inference-time reasoning~\cite{xue2025phyt2v}, or simulation: WonderPlay~\cite{li2025wonderplay} optimizes explicit 4D representations, PSIVG~\cite{foo2026psivg} applies test-time optimization to reconstructed templates, RealWonder~\cite{liu2026realwonder} and PerpetualWonder~\cite{zhan2026perpetualwonder} condition distilled or long-horizon generators on simulation, and PhysGen3D~\cite{chen2025physgen3d} and PhysAnimator~\cite{xie2025physanimator} animate images with simulation. These pipelines rely on model-specific training, test-time optimization, or explicit scene representations, whereas Fluid-Gen-Zero couples simulation with frozen video generators through a training-free, plug-and-play interface.

\section{Method}

\subsection{Overview}

Fluid-Gen-Zero is a training-free image-to-video framework for fluid-object interaction video generation. Given an input image $I_0$, an input prompt $p$, and a frozen pretrained video generator $G_{\theta}$, it produces a realistic video that follows both the prompt and a plausible physical rollout. The framework decouples physical motion from appearance synthesis. The simulator determines what moves and when. The generator determines how the scene looks. As shown in \cref{fig:pipeline}, Fluid-Gen-Zero connects simulation and generation in three stages. \emph{First}, it reconstructs a simulation-ready scene from $I_0$ and $p$. It then rolls out the fluid-object dynamics to obtain a simulation rollout $\mathcal{S}$ (\cref{sec:scene}). \emph{Second}, generation-time planning divides the long rollout into generation clips. It places boundaries in stable intervals and assigns a clip-level prompt to each clip (\cref{sec:planning}). \emph{Finally}, latent-space guidance injects signals from $\mathcal{S}$ into the denoising process. This guides motion in each clip while preserving the pretrained generator's appearance prior (\cref{sec:latent}).

\subsection{Scene Reconstruction and Physics Simulation}
\label{sec:scene}

The simulator and generator need different representations of the same scene. The simulator requires 3D geometry and material parameters, whereas the generator accepts image-space conditions such as masks and optical flow. This stage builds a shared representation so that the two later stages plan and generate clips from the same simulation rollout.

Fluid-Gen-Zero builds this representation in three steps.
\emph{First}, it parses the image and prompt into the intent record $\mathcal{I}$. A rule-based parser maps common motion and material words to simulator parameters, while the VLM resolves image-grounded information such as object category, scene context, and event tags.
\emph{Second}, Fluid-Gen-Zero reconstructs a simulation-ready scene from the input image. It segments $I_0$ into object, fluid, and background regions~\cite{liu2023grounding,ravi2024sam2}, recovers scene geometry as a monocular point map~\cite{moge2}, lifts visible regions into 3D point clouds, and reconstructs the object~\cite{sam3dobjects}. The object pose is refined against the fitted fluid surface so that the visible silhouette and waterline agree.
This gives the initial object point cloud $X_0^o$ and fluid point cloud $X_0^f$, with the background kept as static context.
\emph{Finally}, Fluid-Gen-Zero runs the physics simulator using the reconstructed scene and parsed parameters.
The coupled object-fluid state is advanced by
\begin{equation}
\resizebox{0.9\linewidth}{!}{$
    (X_{\tau+1}^o, X_{\tau+1}^f)
    =
    \Phi_{\Theta_{\mathrm{sim}}}(X_\tau^o, X_\tau^f),
    \quad \tau=0,\ldots,T_{\mathrm{sim}}-1
$}
\end{equation}
where $\tau$ indexes simulation steps, $T_{\mathrm{sim}}$ is the number of steps, $\Phi_{\Theta_{\mathrm{sim}}}$ denotes one simulation step, and $\Theta_{\mathrm{sim}}$ collects simulator parameters parsed from $\mathcal{I}$, such as material properties. We record the state every $N_{\mathrm{rec}}$ steps, which gives 3D particle trajectories with $T$ frames indexed by $t$, and project them into image space. The resulting simulation rollout is
$\mathcal{S}=\{X_{1:T}^{o},X_{1:T}^{f},M_{1:T}^{o},M_{1:T}^{f},F_{1:T-1},R_{1:T}^{\mathrm{sim}},V_{1:T}^{\mathrm{sim}}\}$.
Here, $X^o$ and $X^f$ are the 3D particle trajectories; $M^o$ and $M^f$ are the projected object and fluid masks; $F$ is the projected simulation optical flow; $R^{\mathrm{sim}}$ is the simulation preview; and $V^{\mathrm{sim}}$ is the appearance-preserving warped reference video. Additional details are provided in Appendix~\ref{app:reconstruction_simulation_details} and Appendix~\ref{app:implementation_details}.

\begin{figure*}[t]
    \centering
    \vspace{-0.5em}
\includegraphics[width=\linewidth]{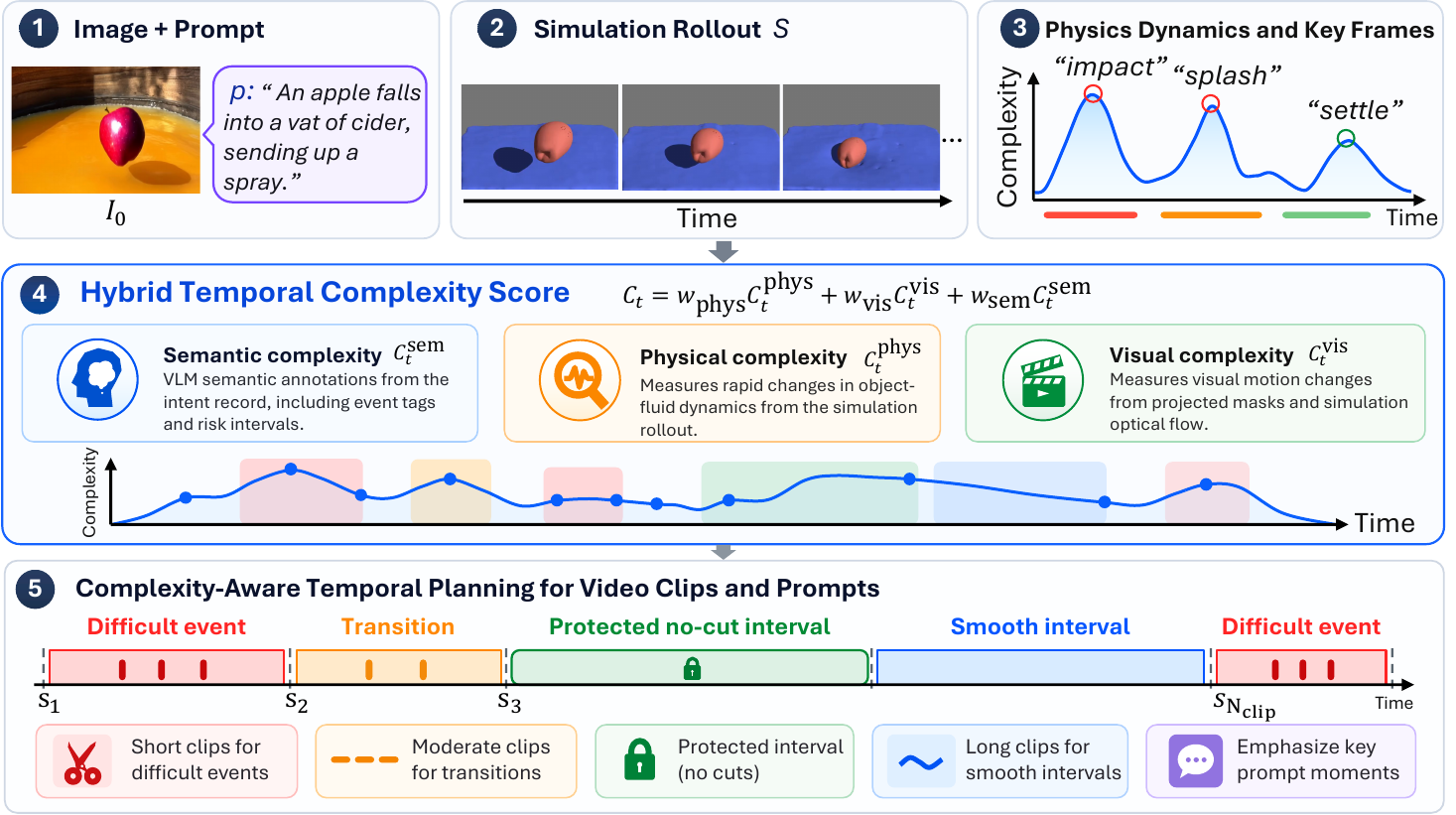}
    \caption{
    \textbf{Generation-time planning for long-horizon video generation.} The planner uses these signals to protect important events, select safe clip boundaries, and refine clip-level prompts.
    }
    \label{fig:planning}
    \vspace{-1.5em}
\end{figure*}
\subsection{Generation-Time Planning for Long-Horizon Video Generation}
\label{sec:planning}

Generation-time planning converts one simulation rollout into a sequence of generation clips that are stitched into a long-horizon video. VFMs support bounded clip lengths, while the difficulty of fluid motion varies substantially over time. Smooth motion allows flexible boundary placement. Abrupt events, such as contact, splashing, and occlusion, should remain within one clip rather than be split across two generations. Accordingly, the planner asks a simple question: \emph{where is it safe to cut?} A safe boundary lies in a temporally stable region. A risky boundary falls inside or near an important physical event. Splitting at such a boundary can introduce motion discontinuity or event inconsistency because the neighboring clips are generated independently. Given the input prompt $p$, intent record $\mathcal{I}$, and simulation rollout $\mathcal{S}$, Fluid-Gen-Zero estimates boundary risk over time. It then produces a clip plan $\Pi=\{(s_i,e_i,p_i)\}_{i=1}^{N_{\mathrm{clip}}}$ containing $N_{\mathrm{clip}}$ clips. Here, $(s_i,e_i)$ specifies the temporal extent of clip $i$, and $p_i$ is its clip-level prompt. \cref{fig:planning} illustrates the workflow. \cref{fig:vis2} shows a representative planning result.

\subsubsection{Hybrid Temporal Complexity Score}

To locate safe clip boundaries, we compute a scalar temporal complexity score $C_t$, which measures how difficult the rollout is around time $t$ and serves as a local boundary-risk signal. We construct $C_t$ by combining three complementary components:
\begin{equation}
C_t
=
w_{\mathrm{phys}} C_t^{\mathrm{phys}}
+
w_{\mathrm{vis}} C_t^{\mathrm{vis}}
+
w_{\mathrm{sem}} C_t^{\mathrm{sem}},
\label{eq:hybrid_complexity}
\end{equation}
where each term captures a different source of boundary risk:

\begin{itemize}[leftmargin=1.4em, itemsep=1pt, topsep=2pt]
\item \textbf{Physical complexity $C_t^{\mathrm{phys}}$.}
Measures rapid changes in the underlying 3D simulated dynamics, including object motion, fluid spread, and object-fluid contact.
\item \textbf{Visual complexity $C_t^{\mathrm{vis}}$.}
Measures how strongly the simulated dynamics change after projection into image space, using signals such as optical-flow magnitude and structural changes in the projected masks.

\item \textbf{Semantic complexity $C_t^{\mathrm{sem}}$.}
Increases around prompt-relevant physical events, such as contact or splashing. Event tags specify which interactions matter, while simulation signals and VLM-estimated importance localize and prioritize them on the rollout timeline.
\end{itemize}
All signals are robustly normalized before aggregation. The weights $w_{\mathrm{phys}},w_{\mathrm{vis}},w_{\mathrm{sem}}$ combine fixed defaults with VLM-predicted priorities and are normalized to sum to one.  The resulting $C_t$ is therefore a temporal complexity score that serves as a local boundary-risk signal: a high value indicates that placing a clip boundary nearby is undesirable. The VLM also identifies semantically important protected event intervals. The planner converts these intervals and high-complexity transitions into the no-cut boundary set $\mathcal{U}$. It then uses the temporal complexity curve and $\mathcal{U}$ to determine clip lengths and evaluate the boundary cost described next. Appendix~\ref{app:planning_details} provides additional details.

\subsubsection{Complexity-Aware Clip Planning}

Generation-time planning converts the temporal complexity curve, the no-cut boundary set $\mathcal{U}$, and input prompt $p$ into the clip plan $\Pi$. The curve determines the preferred clip length. When $C_t$ is low, the planner favors longer clips because each additional boundary requires separately generating and stitching two neighboring clips.
When a complexity peak is approaching, the planner favors a shorter clip and places its boundary before the peak. The complete event can then remain within the next clip and be generated in one forward pass.

Concretely, a candidate clip boundary $b$ is a frame index that can serve as the end frame $e_i$ of the current clip and, if selected, the start frame $s_{i+1}$ of the next clip.
Starting from the current clip start $s$, the planner greedily enumerates candidate end frames supported by the backbone and selects the candidate with the lowest boundary cost. Adjacent clips use the selected boundary as a shared first-frame anchor, and the duplicate anchor frame is removed after stitching. We define the boundary cost as:
\begin{equation}\label{eq:clip_search}
\resizebox{0.9\linewidth}{!}{$
    \mathcal{J}(b)
    =
    \lambda_c \bar{C}_{\mathcal{W}(b)}
    +
    \lambda_l |L(b)-L^{\star}|
    -
    \lambda_e \psi_{\mathrm{evt}}(b)
    +
    \lambda_u \mathds{1}[b\in\mathcal{U}].
$}
\end{equation}
The boundary cost combines four complementary considerations, from soft preferences to a strong no-cut penalty, with $\lambda_c,\lambda_l,\lambda_e,\lambda_u\geq0$:

\begin{itemize}[leftmargin=1.4em, itemsep=1pt, topsep=2pt]
    \item \textbf{Avoid unstable frames.} $\bar{C}_{\mathcal{W}(b)}$ averages complexity over a small neighborhood $\mathcal{W}(b)$, discouraging cuts wherever the rollout is unsettled.
    \item \textbf{Keep the length usable.} $L(b)$ is the clip length induced by ending at $b$, and $L^\star$ is the backbone-supported target length selected from the local complexity level. The term $|L(b)-L^{\star}|$ is the number of frames by which the clip deviates from this target. It prefers boundaries that keep the clip close to $L^\star$, with other terms being equal.
     \item \textbf{Prefer event edges.} $\psi_{\mathrm{evt}}(b)$ rewards boundaries near annotated event edges. Parsed-event endpoints are weighted by event importance, while VLM-suggested cut frames are weighted by VLM confidence.
    \item \textbf{Avoid protected boundaries.} The no-cut boundary set $\mathcal{U}$ contains candidate boundaries on high-complexity transitions or inside protected event intervals. The indicator $\mathds{1}[b\in\mathcal{U}]$ equals $1$ when $b$ is such a protected boundary and $0$ otherwise. 
\end{itemize}

\noindent Once boundaries are fixed, each clip receives a clip-level prompt $p_i$ that preserves the global scene description from $p$ and appends only the physical event local to that interval, such as stable object-fluid contact or detailed splash formation. Full details are in Appendix~\ref{app:planning_details}.

\subsection{Simulation-Guided Latent-Space Guidance}
\label{sec:latent}

Generation-time planning determines where each clip begins and ends, but it does not directly align the generated motion within a clip with the simulation rollout. Simulation-guided latent-space guidance complements the native backbone condition $y_i$
with the latent-control inputs $(V_i^{\mathrm{sim}},M_i^o,M_i^f)$, which are the warped reference video and the object and fluid masks restricted to the frames $[s_i,e_i]$ of clip $i$. Together, the native backbone condition and latent-control inputs form the simulation-derived generation conditions for clip $i$.

\begin{figure}[t]
\centering
\vspace{-0.5em}
\includegraphics[width=\linewidth]{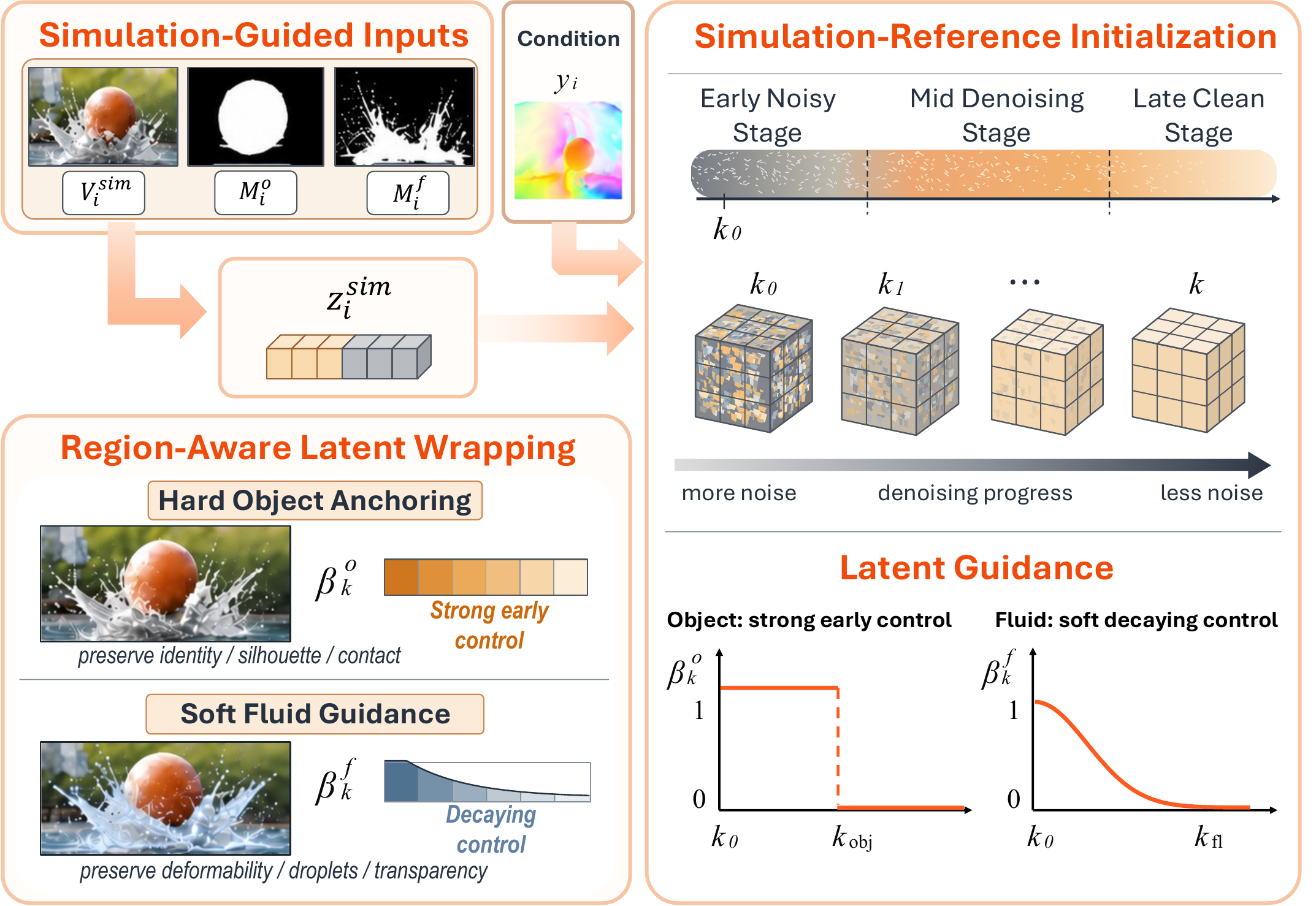}
    \vspace{-1.5em}
\caption{\textbf{Simulation-Guided Latent-Space Guidance.}
The noised reference latent initializes sampling. Object and fluid regions then follow different wrapping schedules during denoising.}
\label{fig:latent_control}
\vspace{-1.5em}
\end{figure}

The guidance has two stages. \emph{First}, Fluid-Gen-Zero initializes sampling from a noised latent encoding of $V_i^{\mathrm{sim}}$, providing an appearance-preserving structural prior from the beginning of denoising. \emph{Second}, it applies region-aware latent wrapping. Object regions receive strong early wrapping to preserve simulated position, shape, identity, and contact structure. Fluid regions receive softer, decaying wrapping that preserves coarse appearance while allowing the generator to synthesize transparency, texture, and other fluid details. \cref{fig:latent_control} illustrates the process. Appendix~\ref{app:latent_control_details} provides additional details.

\begin{figure*}[t!]
    \centering
    \vspace{-1.5em}
    \includegraphics[width=\textwidth]{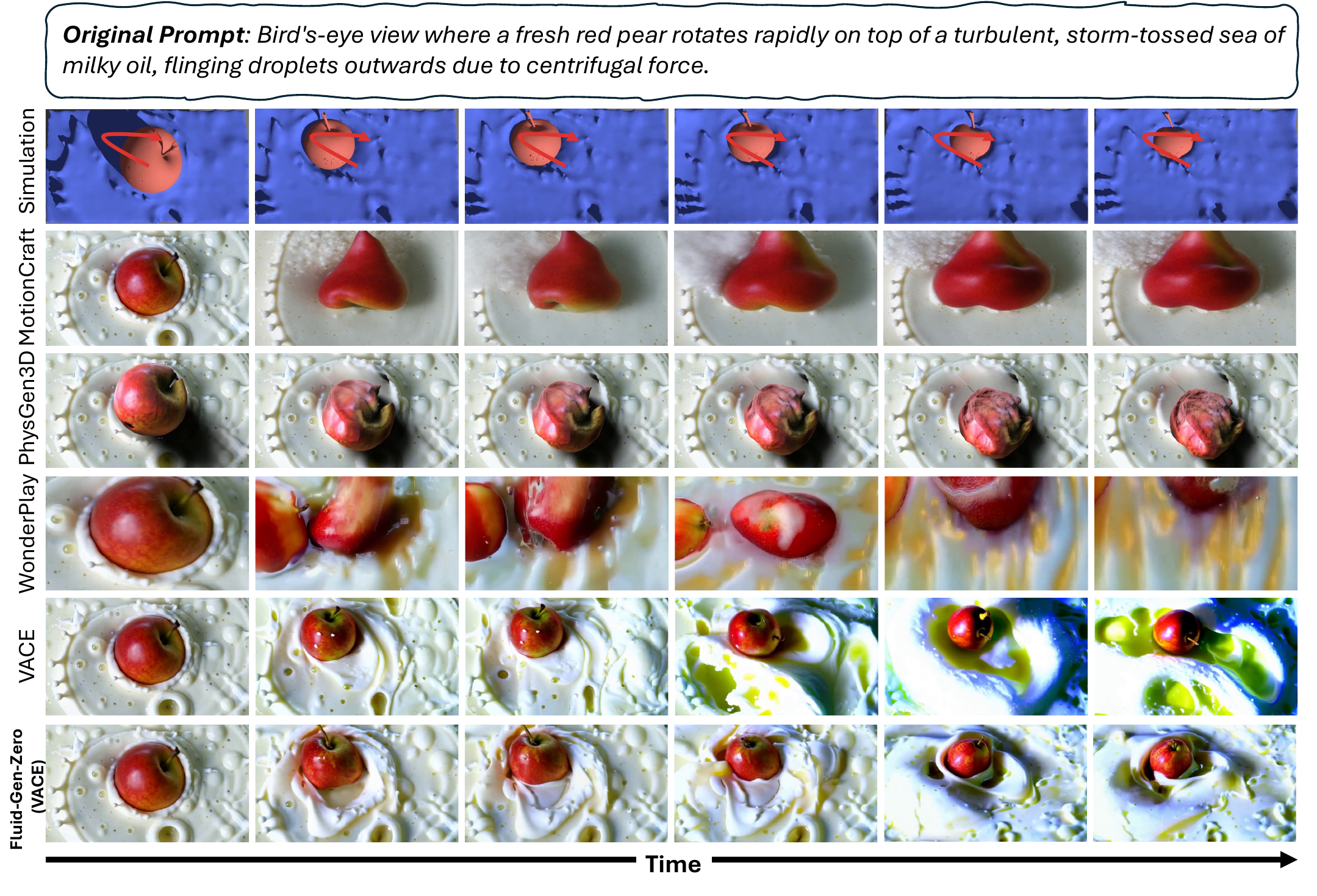}
        \vspace{-2.5em}
       \caption{
\textbf{Qualitative comparison on fluid-object interaction video generation.}
We compare the simulation preview, simulation-based methods (PhysGen3D and WonderPlay), the training-free method MotionCraft, the VACE backbone, and Fluid-Gen-Zero applied to VACE. Each Fluid-Gen-Zero clip uses a refined clip-level prompt derived from the input prompt. }
\vspace{-1.5em}
    \label{fig:vis1}
\end{figure*}

\noindent\textbf{Simulation-Reference Initialization.}
For clarity, we omit the clip index $i$ from the latent variables. For each clip, the warped reference video $V_i^{\mathrm{sim}}$ is encoded by the VAE into a simulation-reference latent $z^{\mathrm{sim}}$, and the masks $M_i^o$ and $M_i^f$ are projected to latent resolution, giving $m^o$ and $m^f$. At denoising step $k$, we construct a noised reference latent
$z^{\mathrm{sim}}_k=\gamma_k z^{\mathrm{sim}}+\sigma_k\epsilon$,
where $\epsilon\sim\mathcal{N}(\mathbf{0},\mathbf{I})$ and $(\gamma_k,\sigma_k)$ are the forward-process coefficients of the backbone's scheduler. For the flow-matching backbones, $\gamma_k=1-\sigma_k$. For Tora, $\gamma_k=\sqrt{\bar{\alpha}_k}$ and $\sigma_k=\sqrt{1-\bar{\alpha}_k}$, where $\bar{\alpha}_k$ is the cumulative signal-retention coefficient provided by its scheduler. We initialize the denoising trajectory at step $k_0$ with $z_{k_0}=z^{\mathrm{sim}}_{k_0}$. This appearance-preserving initialization anchors the object to its simulated configuration while retaining stochasticity for realistic appearance synthesis. It follows the partial-noising principle used in image editing~\cite{meng2021sdedit,singer2025time}, applied here to video latents before region-aware latent wrapping.

\noindent\textbf{Region-Aware Latent Wrapping.}
After initialization, the frozen generator performs its native denoising update at each step, conditioned on the clip-level prompt $p_i$ and native backbone condition $y_i$. We denote the resulting latent by $\tilde{z}_{k+1}$. Fluid-Gen-Zero then injects the noised reference latent $z^{\mathrm{sim}}_{k+1}$ into the object and fluid regions. Unlike $\sigma_k$, which is the scheduler's noise coefficient, $\beta_k^o$ and $\beta_k^f$ determine how strongly the reference is injected during denoising.

\noindent\textbf{Object-region wrapping.}
For object regions, we use \emph{strong early anchoring}:
\begin{equation}
z^o_{k+1}
=
(1-\beta_k^o m^o)\odot\tilde{z}_{k+1}
+
\beta_k^o m^o\odot z^{\mathrm{sim}}_{k+1},
\label{eq:latent_object}
\end{equation}
where $\beta_k^o=1$ for $k_0\leq k<k_{\mathrm{obj}}$ and $\beta_k^o=0$ otherwise. Thus, the object region is fully anchored to the noised reference latent during early denoising and released after $k_{\mathrm{obj}}$, allowing the generator to refine its appearance.

\noindent\textbf{Fluid-region wrapping.}
Starting from the object-controlled latent $z^o_{k+1}$, fluid regions use \emph{soft, decaying control}:
\begin{equation}
z_{k+1}
=
(1-\beta_k^f m^f)\odot z^o_{k+1}
+
\beta_k^f m^f\odot z^{\mathrm{sim}}_{k+1},
\label{eq:latent_fluid}
\end{equation}
where the fluid-control strength gradually decreases over denoising:
$
\beta_k^f
=
\beta^f_{\max}
\frac{k_{\mathrm{fl}}-k}{k_{\mathrm{fl}}-k_0},
k_0\leq k<k_{\mathrm{fl}}.
$
Here, $\beta^f_{\max}$ is the initial fluid-wrapping strength and $k_{\mathrm{fl}}$ marks the end of fluid-region wrapping. For all other denoising steps, $\beta_k^f$ is set to 0. This preserves coarse fluid appearance during early denoising while progressively returning flexibility to the pretrained generator for synthesizing transparency, texture, droplets, and other fine fluid details.

\begin{table*}[t]
\vspace{-0.5em}
\centering
\small
\setlength{\tabcolsep}{2.2pt}

\definecolor{bestbg}{RGB}{255,220,220}
\definecolor{secondbg}{RGB}{255,240,240}

\caption{\textbf{Main results on fluid-object interaction video generation.}
We compare the complete \textbf{Fluid-Gen-Zero} against representative baselines.
\bestmark{} and \secondmark{} results are highlighted.}
\label{tab:main_results}

\resizebox{\linewidth}{!}{
\begin{tabular}{lcccccccccccc}
\toprule

\multirow{2}{*}{Method}
& \multirow{2}{*}{Size}
& \multirow{2}{*}{\makecell{Obj. Traj.\\Err. $\downarrow$}}
& \multirow{2}{*}{\makecell{Fluid\\fEPE $\downarrow$}}
& \multirow{2}{*}{\makecell{Subj.\\Cons. $\uparrow$}}
& \multirow{2}{*}{\makecell{Temp.\\Flick. $\uparrow$}}
& \multirow{2}{*}{\makecell{Aesth.\\Qual. $\uparrow$}}
& \multirow{2}{*}{\makecell{Img.\\Qual. $\uparrow$}}
& \multirow{2}{*}{\makecell{Physical\\Commonsense $\uparrow$}}
& \multicolumn{4}{c}{\textbf{GPT Judge}} \\

\cmidrule(lr){10-13}

&
&
&
&
&
&
&
&
&
\textbf{PhysR} ($\uparrow$)
& \textbf{PhotoR} ($\uparrow$)
& \textbf{Sem} ($\uparrow$)
& \textbf{Overall} ($\uparrow$) \\

\midrule

PhysGen3D
& -
& -
& -
& 0.873
& \cellcolor{bestbg}0.998
& 0.447
& 0.654
& \cellcolor{secondbg}4.791
& 0.260 $\pm$ 0.10
& 0.517 $\pm$ 0.13
& 0.453 $\pm$ 0.21
& 0.410 \\

WonderPlay
& 5B
& -
& -
& 0.731
& 0.991
& 0.362
& 0.435
& 4.558
& 0.298 $\pm$ 0.09
& 0.421 $\pm$ 0.09
& 0.398 $\pm$ 0.15
& 0.372 \\

MotionCraft
& 1B
& 169.62
& 10.659
& 0.836
& 0.977
& 0.438
& 0.598
& \cellcolor{bestbg}4.814
& 0.315 $\pm$ 0.15
& 0.475 $\pm$ 0.19
& 0.475 $\pm$ 0.21
& 0.421 \\

\midrule

VACE
& 1.3B
& 56.02
& 6.445
& 0.733
& 0.978
& 0.461
& 0.590
& 4.116
& 0.385 $\pm$ 0.12
& 0.528 $\pm$ 0.14
& 0.482 $\pm$ 0.19
& 0.465 \\

\textbf{+ Fluid-Gen-Zero}
& 1.3B
& \textbf{41.06}
& \textbf{2.070}
& \textbf{0.782}
& \textbf{0.990}
& \textbf{0.486}
& \textbf{0.642}
& \textbf{4.326}
& \textbf{0.425 $\pm$ 0.11}
& \textbf{0.613 $\pm$ 0.10}
& \textbf{0.578 $\pm$ 0.21}
& \textbf{0.539} \\

\midrule

Tora
& 5B
& 133.20
& 3.950
& 0.804
& 0.980
& 0.487
& 0.656
& 4.512
& 0.477 $\pm$ 0.16
& 0.641 $\pm$ 0.12
& 0.552 $\pm$ 0.23
& 0.556 \\

\textbf{+ Fluid-Gen-Zero}
& 5B
& \cellcolor{secondbg}\textbf{24.61}
& \cellcolor{secondbg}\textbf{0.843}
& \cellcolor{bestbg}\textbf{0.906}
& \cellcolor{secondbg}\textbf{0.995}
& \cellcolor{secondbg}\textbf{0.506}
& \textbf{0.656}
& \textbf{4.628}
& \textbf{0.484 $\pm$ 0.14}
& \cellcolor{bestbg}\textbf{0.691 $\pm$ 0.12}
& \cellcolor{bestbg}\textbf{0.638 $\pm$ 0.22}
& \cellcolor{bestbg}\textbf{0.604} \\

\midrule

WanMove
& 14B
& 92.93
& 5.180
& 0.837
& 0.969
& \cellcolor{bestbg}\textbf{0.528}
& \cellcolor{bestbg}\textbf{0.677}
& 4.209
& \cellcolor{bestbg}\textbf{0.518 $\pm$ 0.15}
& 0.665 $\pm$ 0.12
& 0.596 $\pm$ 0.22
& 0.593 \\

\textbf{+ Fluid-Gen-Zero}
& 14B
& \cellcolor{bestbg}\textbf{23.08}
& \cellcolor{bestbg}\textbf{0.827}
& \cellcolor{secondbg}\textbf{0.904}
& \textbf{0.994}
& 0.500
& \cellcolor{secondbg}0.665
& \textbf{4.302}
& \cellcolor{secondbg}0.486 $\pm$ 0.13
& \cellcolor{secondbg}\textbf{0.680 $\pm$ 0.10}
& \cellcolor{secondbg}\textbf{0.625 $\pm$ 0.22}
& \cellcolor{secondbg}\textbf{0.597} \\

\bottomrule
\end{tabular}}
\vspace{-1em}
\end{table*}

\section{Experiments}

\subsection{Experimental Setup and Evaluation Protocol}
We evaluate Fluid-Gen-Zero on a simulation-driven benchmark of 43 fluid-object interaction samples. Unless otherwise specified, we use 401-frame simulation rollouts for long-horizon generation.
We instantiate the complete Fluid-Gen-Zero framework on three backbones:
VACE~\cite{vace} (1.3B), Tora~\cite{zhang2025tora} (5B), and WanMove~\cite{chu2025wanmove} (14B). We additionally compare against the simulation-based methods PhysGen3D~\cite{chen2025physgen3d} and WonderPlay~\cite{li2025wonderplay}, and the training-free method MotionCraft~\cite{savantaira2024motioncraft}. More implementation and benchmark details are provided in Appendix~\ref{app:implementation_details} and~\ref{app:benchmark_scope}.

\noindent\textbf{Metrics.} Simulation alignment is measured by object trajectory error, the mean per-frame L1 distance in pixels between generated and simulated object centroids, and fluid flow endpoint error (fEPE)~\cite{geng2025motion} between simulation and generated-video optical flow over the fluid region, since a single fluid centroid is ill-defined once the fluid spreads. We also report VBench subject consistency, temporal flickering, aesthetic quality, and image quality~\cite{huang2023vbench}, VideoPhy2 physical commonsense~\cite{bansal2026videophy}, a GPT judge~\cite{chen2025physgen3d} for physical realism (PhysR), photorealism (PhotoR), and semantic consistency (Sem), and a human preference study (\cref{sec:user_study}). More results are provided in Appendix~\ref{app:additional_results}.

\begin{figure*}[t]
    \centering
\includegraphics[width=\linewidth]{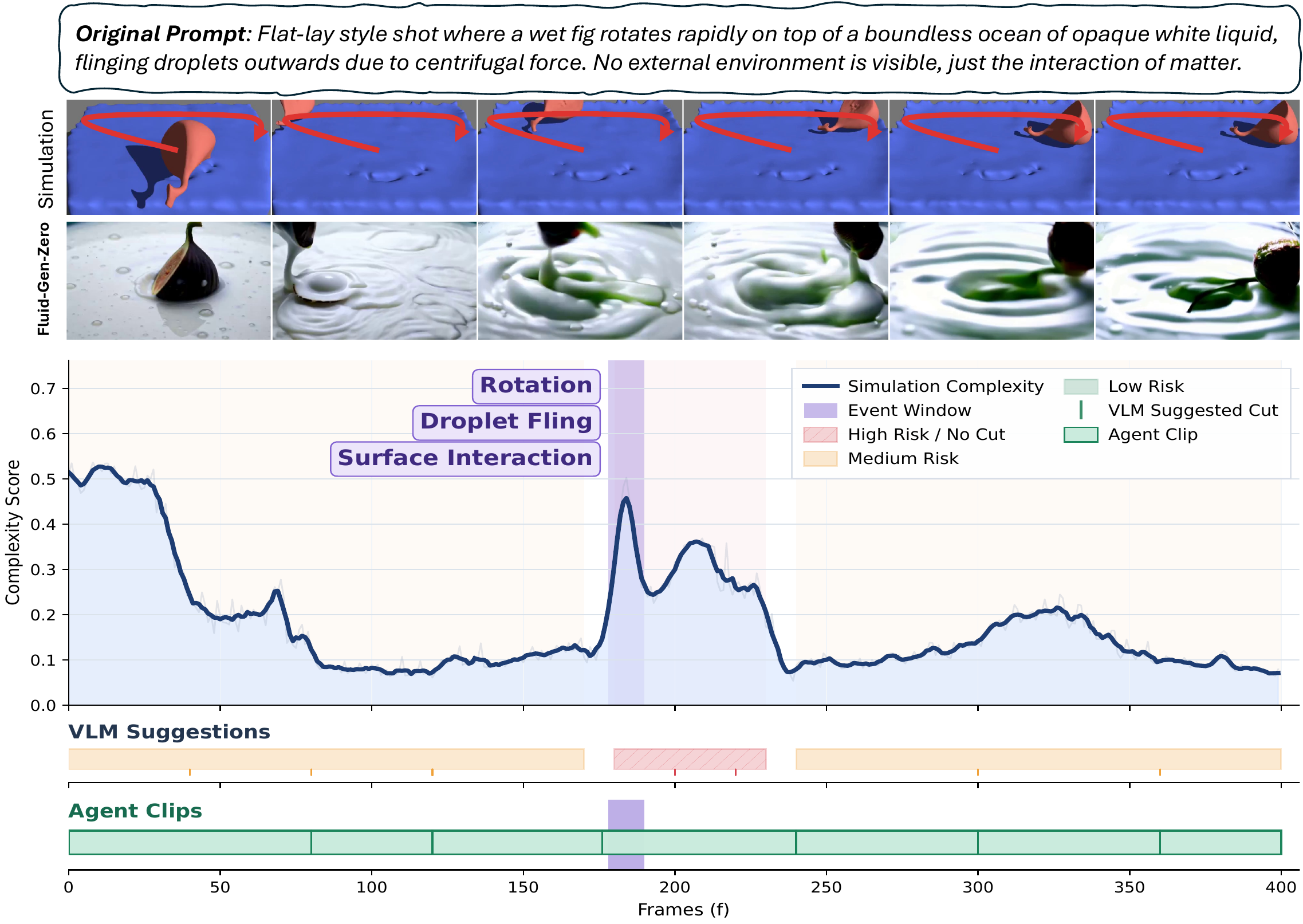}
        \vspace{-2em}
    \caption{\textbf{Generation-time planning and latent-space guidance.}
    The top rows show generated videos from Fluid-Gen-Zero, and the bottom panels visualize the complexity curve.}

    \label{fig:vis2}
    \vspace{-1.5em}
\end{figure*}
\subsection{Main Results: Physics-Aware Fluid-Object Interaction Video Generation}
\noindent\textbf{Simulation-aligned video generation across VFM backbones.}
\Cref{tab:main_results} shows that Fluid-Gen-Zero consistently improves simulation alignment across all three VFM backbones. The largest gains appear on Tora and WanMove, whose backbones are conditioned on sparse motion trajectories: object trajectory error decreases from $133.20$ to $24.61$ and from $92.93$ to $23.08$, and fluid fEPE from $3.950$ to $0.843$ and from $5.180$ to $0.827$, respectively. These results indicate that sparse motion trajectories can specify where an object should move but cannot maintain the complete object-fluid evolution over time. Fluid-Gen-Zero closes much of this gap by anchoring the denoising trajectory to the noised reference latent within each generation clip.

VACE already receives a dense optical-flow condition. Consequently, its baseline object trajectory error is much lower than that of Tora and WanMove, and the relative improvement is more moderate, from $56.02$ to $41.06$. However, dense conditioning alone still does not ensure that the synthesized fluid follows the simulated dynamics: Fluid-Gen-Zero reduces VACE's fluid fEPE from $6.445$ to $2.070$.

Fluid-Gen-Zero is not tied to a particular native control representation: it substantially corrects both object and fluid motion when the backbone receives sparse motion trajectories, and still provides large fluid-motion gains when dense spatial conditioning is already available.

\noindent\textbf{Appearance and semantics are largely preserved.}
Subject consistency and temporal flickering improve on all three backbones, while GPT PhotoR, Sem, and Overall scores also increase for every backbone pair. Fluid-Gen-Zero (Tora) achieves the highest GPT Overall score of $0.604$. On WanMove, aesthetic quality, image quality, and PhysR decrease slightly, reflecting a small trade-off between strong simulation constraints and the pretrained generator's appearance prior.

\noindent\textbf{Comparison with simulation-based and training-free methods.}
PhysGen3D simulates object motion and composites the rendered object over the background but leaves the surrounding fluid largely static. WonderPlay relies on per-scene representation fitting and refinement, making simulation or fitting errors visible in the final appearance. MotionCraft is also training-free and simulation-driven, but performs frame-wise latent warping with an image diffusion model; without a video prior or longer-range constraints, errors accumulate as motion drift and appearance inconsistency. In contrast, Fluid-Gen-Zero retains the pretrained generator's appearance prior while using simulation to constrain both object motion and fluid appearance.

\noindent\textbf{Why physical commonsense scores can be misleading.}
An important observation is that some conventional quality metrics can assign high scores to \textbf{visually stable or superficially plausible motion} even when the underlying dynamics are \textbf{physically implausible}. PhysGen3D achieves the highest temporal flickering score because its simulated object is composited over a largely static background, so most pixels change little over time even though the surrounding fluid remains almost frozen. PhysGen3D and MotionCraft obtain the two highest VideoPhy2 physical commonsense scores despite clear failures in our qualitative results.

This behavior is partly explained by the evaluation protocol. VideoPhy2 evaluates physical commonsense separately from prompt adherence~\cite{bansal2026videophy} and can therefore favor smooth, conservative motion. PhysGen3D inherits smooth object trajectories directly from simulation, while MotionCraft propagates appearance through frame-to-frame latent warping; both mechanisms suppress abrupt motion but do not guarantee realistic long-horizon fluid-object dynamics. The GPT judge, which considers physical realism together with photorealism and semantic consistency, is more sensitive to cases where motion looks locally smooth but the complete generated interaction remains physically or visually implausible. Accordingly, both methods rank among the lowest in GPT Overall, whereas all three Fluid-Gen-Zero variants achieve higher scores. \cref{fig:vis1} shows these differences: PhysGen3D moves the object over an essentially frozen fluid, WonderPlay introduces substantial appearance distortion, and MotionCraft exhibits accumulated drift and flickering.

\noindent\textbf{How generation-time planning adapts to physical complexity.} \cref{fig:vis2} illustrates the planning behavior for a rotating fig that first contacts the fluid surface and then flings droplets outward. The temporal complexity curve rises sharply around the main interaction at approximately frames $180$--$200$. The planner marks this span as a protected event interval and moves the clip boundary to a safer region. It uses longer clips in the lower-complexity portions before and after the event. This result shows how generation-time planning adapts clip boundaries to the temporal structure of the simulation rollout instead of uniformly segmenting it.

\subsection{User Study}
\label{sec:user_study}

We conduct a blind two-alternative forced-choice (2AFC) user study (Appendix~\ref{app:user_study_details}) with same-backbone comparisons between each pretrained generator and its Fluid-Gen-Zero variant, and comparisons against the simulation-based methods PhysGen3D and WonderPlay. Participants judge simulation alignment, temporal consistency, and overall preference. As shown in \cref{tab:user_study}, Fluid-Gen-Zero is preferred in all five comparisons, with $55.1\%$--$74.4\%$ overall preference in the same-backbone setting and $94.2\%$ and $90.4\%$ against PhysGen3D and WonderPlay.

\begin{table}[t]
\vspace{-0.5em}
\centering
\small
\setlength{\tabcolsep}{3.5pt}
\caption{
\textbf{User preference study.}
We conduct blind 2AFC comparisons.
Values greater than 50\% indicate preference for Fluid-Gen-Zero.
}
\vspace{-0.5em}
\label{tab:user_study}
\begin{tabular}{lccc}
\toprule
Comparison
& \makecell{Sim.\\Align. $\uparrow$}
& \makecell{Temp.\\Cons. $\uparrow$}
& \makecell{Overall\\Pref. $\uparrow$} \\
\midrule
\multicolumn{4}{l}{\textit{Same-backbone comparison}} \\
Ours (WanMove) vs.\ WanMove
& 55.1\% & 60.3\% & 55.1\% \\
Ours (Tora) vs.\ Tora
& 69.2\% & 69.2\% & 74.4\% \\
Ours (VACE) vs.\ VACE
& 66.7\% & 69.2\% & 69.2\% \\
\midrule
\multicolumn{4}{l}{\textit{Simulation-based methods}} \\
Fluid-Gen-Zero vs.\ PhysGen3D
& 95.2\% & 85.6\% & 94.2\% \\
Fluid-Gen-Zero vs.\ WonderPlay
& 87.5\% & 92.3\% & 90.4\% \\
\bottomrule
\end{tabular}
\vspace{-2em}
\end{table}

\section{Conclusion}
We presented \textbf{Fluid-Gen-Zero}, a training-free framework that uses simulation as a structural motion prior for frozen video generators through generation-time planning and region-aware latent wrapping, improving simulation alignment, temporal consistency, and human preference across backbones without updating generator parameters. \textbf{Limitations.} Guidance quality depends on reconstruction and simulation accuracy (Appendix~\ref{app:failure_case}); projected masks and optical flow underspecify fine-grained fluid-solid coupling; and long rollouts may still accumulate appearance or camera drift across clips.

\bibliographystyle{ieeenat_fullname}
\bibliography{main}

\clearpage
\setcounter{page}{1}
\maketitlesupplementary
\setcounter{section}{0}
\renewcommand*{\theHsection}{supp.\arabic{section}}
\renewcommand*{\theHsubsection}{supp.\arabic{section}.\arabic{subsection}}

\section{Scene Reconstruction and Physics Simulation Details}
\label{app:reconstruction_simulation_details}
This appendix details the path from a single image to the simulation-derived generation conditions in four stages: \textbf{(i)} parse the input prompt into an intent record, \textbf{(ii)} decompose the image and lift it to 3D, \textbf{(iii)} pose the object against the fluid, and \textbf{(iv)} compute and project the simulation rollout.

\subsection{Image-to-Simulation Scene Setup}
\noindent\textbf{Structured intent record.}
The simulator requires physical attributes, whereas the generator requires descriptive attributes. The intent record separates these fields according to their downstream use, $\mathcal{I}=\{\mathcal{A},\mathcal{D},\mathcal{E}\}$:

\begin{itemize}[leftmargin=1.4em, itemsep=2pt, topsep=3pt]
    \item $\mathcal{A}$, \textbf{appearance}, is what the \emph{generator} must preserve: object and fluid appearance, material description, lighting, background, camera framing, and style. These fields are reused when building clip-level prompts, so a local refinement preserves the scene's visual identity.
    \item $\mathcal{D}$, \textbf{dynamics}, is what the \emph{simulator} consumes: fluid type, material parameters, external-force magnitude and direction, flow direction and speed, object material and elasticity, and initial motion. Material words map to parameter ranges (``water'', ``oil'', and ``mud'' to different viscosity and density), motion words to initial conditions (``pour'', ``splash'', ``float'', ``sink''). These values become the simulator parameters $\Theta_{\mathrm{sim}}$.
    \item $\mathcal{E}$, \textbf{events}, is what the \emph{planner} needs: each entry carries a type, an importance score, a localization confidence, and the simulation signals that localize it on the timeline. Contact is localized by object-fluid distance and motion magnitude, splash by fluid spread and mask fragmentation, swirl by curl-like flow, and occlusion by mask overlap and silhouette change. These events determine the semantic complexity term and protected event intervals.
\end{itemize}

We use two complementary components here. A \textbf{rule-based parser} maps material and motion words to valid ranges and provides each numerical simulation parameter: viscosity, density, force magnitude, and flow speed. The \textbf{VLM} resolves image-grounded attributes: object identity, visible fluid region, material type, scene context, and event importance.

\noindent\textbf{Scene decomposition.}
The input image is treated as the first video frame and split into object, fluid, and static background regions, yielding object and fluid masks plus a static background reference.
\noindent\textbf{Region extraction.}
Grounding DINO~\cite{liu2023grounding} proposes text-grounded boxes and SAM2~\cite{ravi2024sam2} converts them into pixel masks. Two aspects are specific to fluid scenes:

\begin{itemize}[leftmargin=1.4em, itemsep=1pt, topsep=3pt]
    \item Grounding DINO uses the fluid category named in the intent record, while the object is queried separately.
    \item Where the two masks overlap, the object mask is subtracted from the fluid mask, producing disjoint object and fluid regions.
\end{itemize}

\noindent Scene geometry then comes from a monocular point map~\cite{moge2}, which returns per-pixel 3D positions together with the camera model, and the fluid and background points are read directly from it. Because every later stage consumes \emph{the same} point map, the object, the fluid surface, and the camera share one geometry.

\subsection{Reconstruction and Registration}

We pass the segmented crop to an image-conditioned 3D generation model~\cite{sam3dobjects} along with the point map from the previous step.

\noindent\textbf{Pose refinement against the waterline.}
The initial pose does not account for fluid occlusion. Because the fluid can hide the submerged portion of the object, the observed mask stops at the waterline. Scoring the full rendered silhouette against this mask would penalize a correctly predicted submerged region and displace the object upward. We therefore clip the mesh by the fitted fluid plane and score only the region that should be visible, combining silhouette agreement above the surface with a depth residual. The refined mesh pose is anchored by the point-map observations, and the posed object is then discretized into the initial object particle state $X_0^o$ for simulation.

\noindent\textbf{Fluid and background.}

\begin{itemize}[leftmargin=1.4em, itemsep=1pt, topsep=3pt]
    \item The \textbf{fluid region} is lifted directly from the point map, rescaled to metric units by the estimated object size, and discretized into the initial fluid particle state $X_0^f$ for simulation.
    \item The \textbf{background} is lifted in the same coordinate system but kept only as static visual context and as geometric evidence for support-plane and camera estimation. It is never simulated.
\end{itemize}

\subsection{Simulation Rollout and Simulation-Derived Generation Conditions}
\noindent\textbf{Simulator configuration.}
The reconstructed state and intent record together instantiate the simulator. Material words such as rigid, metallic, glass, rubber, or soft select priors for density, viscosity, and elasticity, and every value is clamped to a plausible range. The configuration is
\begin{equation}
    \Theta_{\mathrm{sim}}
    =
    \{\Delta t,N_{\mathrm{sub}},\mathbf{g},E^o,\nu^o,\rho^f,\mu^f\},
\end{equation}
where $\Delta t$ is the time step, $N_{\mathrm{sub}}$ the number of substeps, $\mathbf{g}$ gravity, $(E^o,\nu^o)$ the object material parameters, and $(\rho^f,\mu^f)$ the fluid material parameters. When requested by the input prompt, the rule-based parser also specifies an external force magnitude and direction.

\noindent\textbf{Anchoring the world to the fluid.}
Before rollout we fit the fluid surface and adopt that plane as the world frame: the surface is the ground plane, its normal is up, and gravity points along the negative normal by construction. Anchoring to the fluid rather than the camera removes the orientation ambiguity of an image-space estimate and places reconstruction and simulation in one coordinate frame. Four further setup choices follow:

\begin{itemize}[leftmargin=1.4em, itemsep=1pt, topsep=3pt]
    \item The \textbf{fluid} is represented by a liquid MPM material filling the domain below the surface, with the object's volume removed so that no fluid is trapped inside the object.
    \item The \textbf{object} is built from the reconstructed mesh, which retains the geometry hidden from the camera and therefore the volume it actually displaces. A rigid object uses the mesh directly; an elastic object fills it with MPM particles.
    \item \textbf{Domain walls} sit outside the visible fluid rather than on its bounding box, so waves are not reflected back into the scene at the frame edge.
    \item \textbf{Force}, when the prompt requests it, is applied along the recovered tangent direction of the surface.
\end{itemize}

\noindent\textbf{Rollout and projection.}
The simulator advances the coupled object-fluid state under gravity, contact, material response, and an optional external force. The resulting simulation rollout contains 3D particle trajectories for the object and fluid. The state is recorded every $N_{\mathrm{rec}}$ simulation steps to form the $T$-frame trajectories, which are then projected into image space. The projected particles define the object mask $M_{1:T}^{o}$ and fluid mask $M_{1:T}^{f}$, consecutive positions define the projected simulation optical flow $F_{1:T-1}$, and a lightweight simulation preview $R_{1:T}^{\mathrm{sim}}$ presents the rollout to the VLM.

\noindent\textbf{Appearance-preserving warped reference video.}
The simulation preview $R_{1:T}^{\mathrm{sim}}$ is physically informative but not photorealistic, so encoding it directly would inject rendering artifacts. We therefore construct an appearance-preserving warped reference video $V_{1:T}^{\mathrm{sim}}$ that combines simulated motion with the appearance of the input image:

\begin{enumerate}[leftmargin=1.4em, itemsep=1pt, topsep=3pt]
    \item reuse the object mask from scene decomposition and remove the object from the first frame by mask-guided inpainting, giving a clean static background;
    \item translate the original object layer by the centroid displacement of the simulated object masks;
    \item alpha-composite it over the inpainted background at every frame.
\end{enumerate}

\noindent This yields the warped reference video $V_{1:T}^{\mathrm{sim}}$ and its masks for the region-aware latent wrapping described in \cref{sec:latent}.

\noindent\textbf{Simulation-derived generation conditions.}
The same simulation rollout $\mathcal{S}$ serves both planning and guidance. For each planned interval $[s_i,e_i]$, it is converted into two groups based on how they reach the generator:

\begin{itemize}[leftmargin=1.4em, itemsep=1pt, topsep=3pt]
    \item the \textbf{native backbone condition} $y_i$, built from the input image, projected simulation optical flow, and projected masks, entering through the conditioning path exposed by each backbone;
    \item the \textbf{latent-control inputs} $(V_i^{\mathrm{sim}},M_i^o,M_i^f)$, consisting of the warped reference video, object mask, and fluid mask for the same interval.
\end{itemize}

\section{Generation-Time Planning Details}
\label{app:planning_details}

\subsection{Temporal Complexity Computation}

Generation-time planning combines measured simulation signals with semantic annotations from the VLM. Given the intent record $\mathcal{I}$ and simulation rollout $\mathcal{S}$, it produces the clip plan $\Pi$, clip-level prompts, event tags, protected event intervals, and simulation-derived generation conditions. The planner first computes an initial complexity curve without VLM risk-interval annotations. The VLM observes this curve together with the simulation preview and returns semantic annotations. The planner then computes the final temporal complexity curve, constructs the no-cut boundary set $\mathcal{U}$, and selects the clip boundaries. Boundary construction proceeds in 6 steps; prompt planning follows once the boundaries are fixed:
\begin{enumerate}[leftmargin=1.4em, itemsep=1pt, topsep=3pt]
    \item \textbf{Compute raw signals.} Measure physical changes from the 3D particle trajectories and visual changes from projected simulation optical flow and masks.
    \item \textbf{Normalize.} Map all signals to a common $[0,1]$ scale.
    \item \textbf{Compute physical and visual components.} Aggregate the normalized simulation signals by family.
    \item \textbf{Add semantic annotations.} Localize prompt-parsed events, form the initial complexity curve without VLM risk-interval annotations, and use the VLM to identify risk intervals with no-cut flags and suggested cut frames.
    \item \textbf{Fuse complexity and protect events.} Compute the final temporal complexity score $C_t$ and construct $\mathcal{U}$.
    \item \textbf{Select boundaries.} Evaluate the boundary cost in \cref{eq:clip_search} over backbone-valid candidates.
\end{enumerate}

\noindent\textbf{Steps 1--2: signal computation and normalization.}
Each raw signal $\phi_m(t)$ is a scalar measured at transition $t\rightarrow t+1$. The signals have different units. We normalize each one to $[0,1]$ by mapping its $5$th percentile to $0$ and its $95$th percentile to $1$, then clipping values outside this range. We denote the normalized signal by $\hat\phi_m(t)$.

\noindent\textbf{Step 3: physical and visual components.}
The physical and visual components identify rapid state changes that make nearby clip boundaries risky. Their signals form two families:

\begin{itemize}[leftmargin=1.4em, itemsep=1pt, topsep=3pt]
    \item $\mathcal{Q}_{\mathrm{phys}}$, read from the \textbf{3D particle trajectories} (speeds, accelerations, how far the particles spread, how close the fluid comes to the object);
    \item $\mathcal{Q}_{\mathrm{vis}}$, read from the \textbf{projected simulation optical flow and masks} (flow magnitude, curl, mask fragmentation, silhouette change).
\end{itemize}

\noindent Each component, $C_t^{\mathrm{phys}}$ or $C_t^{\mathrm{vis}}$, averages the available $\hat\phi_m(t)$ of its family at transition $t$. \cref{tab:physical_visual_metric_components} defines every signal. Three quantities recur in these definitions. For a state $r\in\{o,f\}$ with particle positions $\mathbf{x}_{t,n}^r$, they are:

\begin{itemize}[leftmargin=1.4em, itemsep=1pt, topsep=3pt]
    \item the centroid $\mathbf{c}_t^r$, the mean particle position;
    \item the spread $\mathrm{spr}_t^r$, the mean distance from a particle to that centroid, which grows as the fluid disperses;
    \item the contact distance $d$, from each subsampled fluid particle to its nearest object particle, compared against a scene-adaptive threshold $\delta_{\mathrm{cnt}}$ set from the initial bounding-box diagonal.
\end{itemize}

\begin{table*}[t]
    \centering
    \caption{\textbf{Signals used for temporal complexity computation.}
The physical signals in $\mathcal{Q}_{\mathrm{phys}}$ are aggregated to compute $C_t^{\mathrm{phys}}$, while the optical-flow and projected-mask signals in $\mathcal{Q}_{\mathrm{vis}}$ are aggregated to compute $C_t^{\mathrm{vis}}$.}
    \label{tab:physical_visual_metric_components}
    \vspace{2mm}
    \small
    \begin{tabular}{p{0.10\linewidth}p{0.26\linewidth}p{0.54\linewidth}}
        \toprule
        Symbol & Signal & Definition at transition $t\rightarrow t+1$ \\
        \midrule
        \multicolumn{3}{p{0.92\linewidth}}{\textit{\textbf{Physical family $\mathcal{Q}_{\mathrm{phys}}$: 3D particle trajectories, computed per state $r\in\{o,f\}$}}} \\
        $\phi^r_{\mathrm{cspd}}$ & Centroid speed & $\|\mathbf{c}_{t+1}^r-\mathbf{c}_t^r\|_2$. \\
        $\phi^r_{\mathrm{vmean}}$ & Mean particle speed & Mean of $\|\mathbf{x}_{t+1,n}^r-\mathbf{x}_{t,n}^r\|_2$ over particles $n$. \\
        $\phi^r_{\mathrm{vvar}}$ & Particle-speed variance & Variance of the same displacement magnitudes over $n$. \\
        $\phi^r_{\mathrm{p95}}$ & High-percentile displacement & $95$th percentile of the displacement magnitudes over $n$. \\
        $\phi^r_{\mathrm{acc}}$ & Particle acceleration & Mean magnitude of the finite difference of displacement vectors. \\
        $\phi^r_{\mathrm{spr}}$ & Spread change & $|\mathrm{spr}_{t+1}^r-\mathrm{spr}_t^r|$. \\
        $\phi^r_{\mathrm{ent}}$ & Occupancy-entropy change & Absolute change of particle occupancy entropy on a coarse spatial grid. \\
        \addlinespace[2pt]
        \multicolumn{3}{p{0.92\linewidth}}{\textit{Contact signals, defined only for object-fluid interaction scenes}} \\
        $\phi_{\mathrm{cnt}}$ & Contact intensity & Fraction of sampled fluid particles with nearest object distance $d\leq\delta_{\mathrm{cnt}}$. \\
        $\phi_{\mathrm{cls}}$ & Contact closeness & Mean of $\exp(-d/\delta_{\mathrm{cnt}})$ over sampled fluid-to-object nearest distances. \\
        $\phi_{\Delta\mathrm{cnt}}$ & Contact change & Absolute frame-to-frame change of $\phi_{\mathrm{cnt}}$. \\
        \midrule
        \multicolumn{3}{p{0.92\linewidth}}{\textit{\textbf{Visual family $\mathcal{Q}_{\mathrm{vis}}$, part 1: projected simulation optical flow $F_t=(F_t^x,F_t^y)$}}} \\
        $\phi_{\mathrm{fmean}}$ & Mean flow magnitude & Mean of $\|F_t\|$ over image pixels. \\
        $\phi_{\mathrm{fvar}}$ & Flow-magnitude variance & Variance of $\|F_t\|$ over image pixels. \\
        $\phi_{\mathrm{fp95}}$ & High-percentile flow magnitude & $95$th percentile of $\|F_t\|$ over image pixels. \\
        $\phi_{\mathrm{act}}$ & Active-motion ratio & Fraction of pixels with $\|F_t\|$ above a small motion threshold. \\
        $\phi_{\mathrm{sal}}$ & Motion saliency concentration & Fraction of total flow magnitude held by the top $10\%$ highest-motion pixels. \\
        $\phi_{\mathrm{disc}}$ & Flow discontinuity & Mean spatial gradient magnitude of the two flow channels. \\
        $\phi_{\mathrm{curl}}$ & Flow curl & Mean of $|\partial_x F_t^y-\partial_y F_t^x|$. \\
        $\phi_{\mathrm{div}}$ & Flow divergence & Mean of $|\partial_x F_t^x+\partial_y F_t^y|$. \\
        \midrule
        \multicolumn{3}{p{0.92\linewidth}}{\textit{\textbf{Visual family $\mathcal{Q}_{\mathrm{vis}}$, part 2: projected object and fluid masks}}} \\
        $\phi_{\Delta\mathrm{area}}$ & Mask-area change & Absolute change of the foreground-pixel ratio. \\
        $\phi_{\mathrm{bnd}}$ & Boundary density & Fraction of image pixels on the eroded-mask boundary. \\
        $\phi_{\mathrm{frag}}$ & Fragmentation & Number of connected foreground components under 8-connectivity. \\
        $\phi_{\Delta\mathrm{frag}}$ & Fragmentation change & Absolute change of the connected-component count. \\
        $\phi_{\mathrm{sil}}$ & Silhouette change & XOR-over-union ratio between consecutive masks. \\
        \bottomrule
    \end{tabular}
\end{table*}

\noindent\textbf{Step 4: VLM annotations and semantic component.}
Physical and visual signals identify where the rollout changes, but not which changes are important to the input prompt. Each event $j$ parsed from the prompt carries an alignment set $\mathcal{Q}^{\mathrm{evt}}_j$, listed in \cref{tab:event_metric_mapping}, whose signals locate the peak of that event on the simulation timeline. A Gaussian window around the peak forms an event curve weighted by event importance and localization confidence. The resulting event curves, localized from simulation signals, form the initial complexity curve together with the physical and visual components.

The VLM observes the prompt, sampled frames from the simulation preview, optical-flow visualizations, object and fluid masks, and this initial curve. It returns semantic risk intervals. Each risk interval contains frame endpoints, a categorical risk label (low, medium, or high), a no-cut flag, suggested cut frames, and a confidence value $\xi\in[0,1]$. We assign the interval a score of $0.3\xi$, $0.6\xi$, or $0.9\xi$ for low, medium, or high risk, respectively, and use the maximum score where intervals overlap. The resulting VLM risk curve is averaged with the event curves localized from simulation signals to produce $C_t^{\mathrm{sem}}$.

\begin{table*}[t]
    \centering
    \caption{Alignment sets $\mathcal{Q}^{\mathrm{evt}}_j$ used for semantic complexity.
    The last row is the fallback for events whose type is not recognized.}
    \label{tab:event_metric_mapping}
    \vspace{2mm}
    \small
    \begin{tabular}{p{0.18\linewidth}p{0.30\linewidth}p{0.40\linewidth}}
        \toprule
        Event type & Alignment set $\mathcal{Q}^{\mathrm{evt}}_j$ & Signal names \\
        \midrule
        impact / contact & $\{\phi_{\mathrm{cnt}},\,\phi_{\mathrm{fp95}},\,\phi_{\mathrm{sil}}\}$ & contact intensity, high-percentile flow, silhouette change \\
        splash & $\{\phi^{f}_{\mathrm{p95}},\,\phi_{\mathrm{disc}},\,\phi_{\mathrm{frag}}\}$ & high-percentile fluid displacement, flow discontinuity, fragmentation \\
        split / merge & $\{\phi_{\mathrm{frag}},\,\phi^{f}_{\mathrm{spr}},\,\phi_{\mathrm{disc}}\}$ & fragmentation, fluid spread change, flow discontinuity \\
        swirl & $\{\phi_{\mathrm{curl}},\,\phi_{\mathrm{fvar}},\,\phi^{f}_{\mathrm{vvar}}\}$ & flow curl, flow-magnitude variance, fluid particle-speed variance \\
        occlusion & $\{\phi_{\mathrm{bnd}},\,\phi_{\mathrm{sil}},\,\phi_{\Delta\mathrm{cnt}}\}$ & boundary density, silhouette change, contact change \\
        material detail & $\{\phi_{\mathrm{fmean}},\,\phi_{\mathrm{bnd}}\}$ & mean flow magnitude, boundary density \\
        default interaction & $\{\phi_{\mathrm{cnt}},\,\phi_{\mathrm{fmean}},\,\phi^{f}_{\mathrm{spr}}\}$ & contact intensity, mean flow magnitude, fluid spread change \\
        \bottomrule
    \end{tabular}
\end{table*}

\noindent\textbf{Step 5: final complexity curve and protected boundaries.}
The three components $C_t^{\mathrm{phys}}$, $C_t^{\mathrm{vis}}$, and $C_t^{\mathrm{sem}}$ form the complexity curve in \cref{eq:hybrid_complexity}. Their weights are computed as follows:
\begin{enumerate}[leftmargin=1.4em, itemsep=1pt, topsep=3pt]
    \item During intent parsing, the VLM uses the prompt semantics to produce one non-negative priority for each family.
    \item We normalize these priorities to sum to one. They control only the relative contribution of the three families.
    \item We average the normalized priorities with the fixed prior $\bar{w}=(0.40,0.40,0.20)$ over the physical, visual, and semantic families, using equal weights, then renormalize the result.
\end{enumerate}

\noindent The resulting sequence $\{C_t\}_{t=1}^{T-1}$ is the final temporal complexity curve used for boundary selection. The planner combines this curve with the semantic annotations to construct the no-cut boundary set $\mathcal{U}$. A candidate boundary $b$ is added to $\mathcal{U}$ when the preceding transition satisfies $C_{b-1}\geq0.72$, when $b$ lies strictly inside a parsed event with importance at least $0.75$, or when it lies strictly inside a VLM interval marked as no-cut. We refer to the latter two types of intervals as protected event intervals.

\subsection{Boundary Selection}

\noindent\textbf{Step 6: selecting safe boundaries.}
The score $C_t$ is defined on the transition from frame $t$ to frame $t+1$. A candidate boundary $b$ is a frame index. Adjacent clips share this anchor frame, and the duplicate is removed after stitching. Starting from the current frame $s$, the planner searches from left to right:

\begin{enumerate}[leftmargin=1.4em, itemsep=1pt, topsep=3pt]
    \item \textbf{Pick a target length.} Inspect $C_t$ within the longest feasible horizon. A high peak selects the short target, a medium average selects the default target, and a stable interval selects the long target. We use $(L_{\min},L_{\mathrm{def}},L_{\max})=(13,41,81)$ frames with medium threshold $0.45$ and the same high threshold $0.72$ used for $\mathcal{U}$. This selects $L^\star\in\{L_{\min},L_{\mathrm{def}},L_{\max}\}$.
    \item \textbf{List the candidates.} The backbone accepts only clip lengths $L$ with $L\equiv1 \pmod{4}$. Each length $L$ gives one boundary $b=s+L-1$.
    \item \textbf{Evaluate boundary costs.} Compute \cref{eq:clip_search} for every candidate and select the lowest-cost boundary. We set $\mathcal{W}(b)=\{b-2,b-1,b,b+1\}\cap\{1,\ldots,T-1\}$, $\lambda_c=1$, $\lambda_l=0.35/L_{\max}$, $\lambda_e=1$, and $\lambda_u=10$. The event bonus $\psi_{\mathrm{evt}}(b)$ is $0.25$ times the maximum available event-edge score within two frames of $b$: parsed-event endpoints use event importance, while VLM-suggested cut frames use VLM confidence. Taking the maximum prevents overlapping annotations from accumulating; the bonus is zero when no event edge is nearby. A candidate receives the no-cut penalty $\lambda_u$ when it lies on a high-complexity transition or inside a protected event interval.

    \item \textbf{Advance.} Emit the clip $[s,b]$, set $s\leftarrow b$, and repeat until the remaining frames fit in one final clip.
\end{enumerate}

\subsection{Prompt Planning and Stitching}
Once clip boundaries are fixed, a separate VLM call produces optional clip-level prompt refinements. For each selected clip, the planner starts from the input prompt $p$, keeps the global scene description, and uses the semantic events and dominant risk tags within the clip to request a local refinement.
The resulting clip-level prompt $p_i$ preserves subject identity, material, background, lighting, and camera framing, then appends local physical emphasis.
For example, a contact-heavy interval emphasizes boundary consistency and stable contact geometry, a splash-heavy interval emphasizes droplets and thin fluid sheets, and a topology-change interval emphasizes coherent splitting or merging.
The final metadata for each clip contains $(s_i,e_i)$, the clip-level prompt $p_i$, a risk summary, event tags, the relevant entries of $\mathcal{U}$, and the simulation-derived generation conditions passed to latent-space guidance.
For long-horizon generation, the last generated frame of clip $i$ is reused as the anchor frame for clip $i+1$ when the backbone supports image anchoring. After generation, the duplicated anchor frame is removed before stitching.
This anchor frame reduces clip-boundary drift without requiring backbone fine-tuning.

\section{Simulation-Guided Latent-Space Guidance Details}
\label{app:latent_control_details}
The object/fluid asymmetry of \cref{sec:latent} appears in two places: where latent wrapping is applied and how long it persists. As in \cref{sec:latent}, we omit the clip index $i$ from latent variables.

\subsection{Latent Mask Projection}
For each clip, we encode the warped reference video $V_i^{\mathrm{sim}}$ as the simulation-reference latent $z^{\mathrm{sim}}$ and project the object mask $M_i^o$ and fluid mask $M_i^f$ to latent resolution, giving $m^o$ and $m^f$. The two masks reflect the different wrapping requirements of objects and fluids:
\begin{itemize}[leftmargin=1.4em, itemsep=2pt, topsep=3pt]
\item The \textbf{object mask} $m^o$ remains binary to preserve object geometry, identity, and contact structure under strong early wrapping.
\item The \textbf{fluid mask} $m^f$ is Gaussian-smoothed to avoid sharp latent discontinuities near deformable fluid boundaries.
\end{itemize}
Both masks are broadcast over latent channels during wrapping.

\subsection{Reference Initialization and Wrapping Schedules}
\noindent\textbf{Initialization.}
Initialization fixes how strongly the reference constrains the \emph{early} denoising trajectory. Pure noise leaves the early path unconstrained, so later control steps must correct a layout that is already wrong. The clean reference over-constrains appearance and carries proxy-rendering artifacts into the output. We therefore construct the noised reference latent as
$z^{\mathrm{sim}}_k=\gamma_k z^{\mathrm{sim}}+\sigma_k\epsilon$,
where $\epsilon\sim\mathcal{N}(\mathbf{0},\mathbf{I})$ has the same shape as $z^{\mathrm{sim}}$, and $(\gamma_k,\sigma_k)$ are obtained from the backbone's forward-process scheduler. For the flow-matching backbones, $\gamma_k=1-\sigma_k$. For Tora, $\gamma_k=\sqrt{\bar{\alpha}_k}$ and $\sigma_k=\sqrt{1-\bar{\alpha}_k}$, where $\bar{\alpha}_k$ is the scheduler's cumulative signal-retention coefficient. Sampling starts from $z^{\mathrm{sim}}_{k_0}$, which anchors coarse layout and motion while retaining stochasticity for the generator to synthesize texture, lighting, transparency, and fine detail.

\noindent\textbf{Wrapping schedules.}
At each denoising step, the frozen generator first produces $\tilde{z}_{k+1}$ using the clip-level prompt $p_i$ and native backbone condition $y_i$. Region-aware latent wrapping is then applied sequentially: object-region wrapping produces $z^o_{k+1}$ according to \cref{eq:latent_object}, followed by fluid-region wrapping that produces $z_{k+1}$ according to \cref{eq:latent_fluid}. The two regions use different temporal schedules:

\begin{itemize}[leftmargin=1.4em, itemsep=3pt, topsep=3pt]
\item \textbf{Object: binary, early.} We set $\beta_k^o=1$ for $k_0\leq k<k_{\mathrm{obj}}$ and $\beta_k^o=0$ otherwise. During $[k_0,k_{\mathrm{obj}})$, the noised reference latent replaces the object region of $\tilde{z}_{k+1}$, strongly anchoring object identity, silhouette, and contact geometry while the global structure forms. After $k_{\mathrm{obj}}$, the generator refines the object's appearance without further object-region wrapping.

\item \textbf{Fluid: continuous, decaying.} For $k_0\leq k<k_{\mathrm{fl}}$, we set
$
\beta_k^f
=
\beta^f_{\max}
\frac{k_{\mathrm{fl}}-k}{k_{\mathrm{fl}}-k_0},
$
and $\beta_k^f=0$ otherwise. Applied to the object-wrapped latent $z^o_{k+1}$ in \cref{eq:latent_fluid}, this schedule blends the noised reference latent into the fluid region. Its strength decreases linearly from $\beta^f_{\max}$ at $k_0$ to $0$ at $k_{\mathrm{fl}}$. Stronger early wrapping preserves coarse fluid layout and appearance, while the gradual decay gives the generator increasing freedom to synthesize transparency, texture, droplets, and other fine fluid details.

\end{itemize}

\noindent The binary object schedule prioritizes stable geometry during early denoising, whereas the decaying fluid schedule preserves coarse motion while progressively returning appearance synthesis to the pretrained generator.

\begin{figure*}[t]
    \centering
    \includegraphics[width=\linewidth]{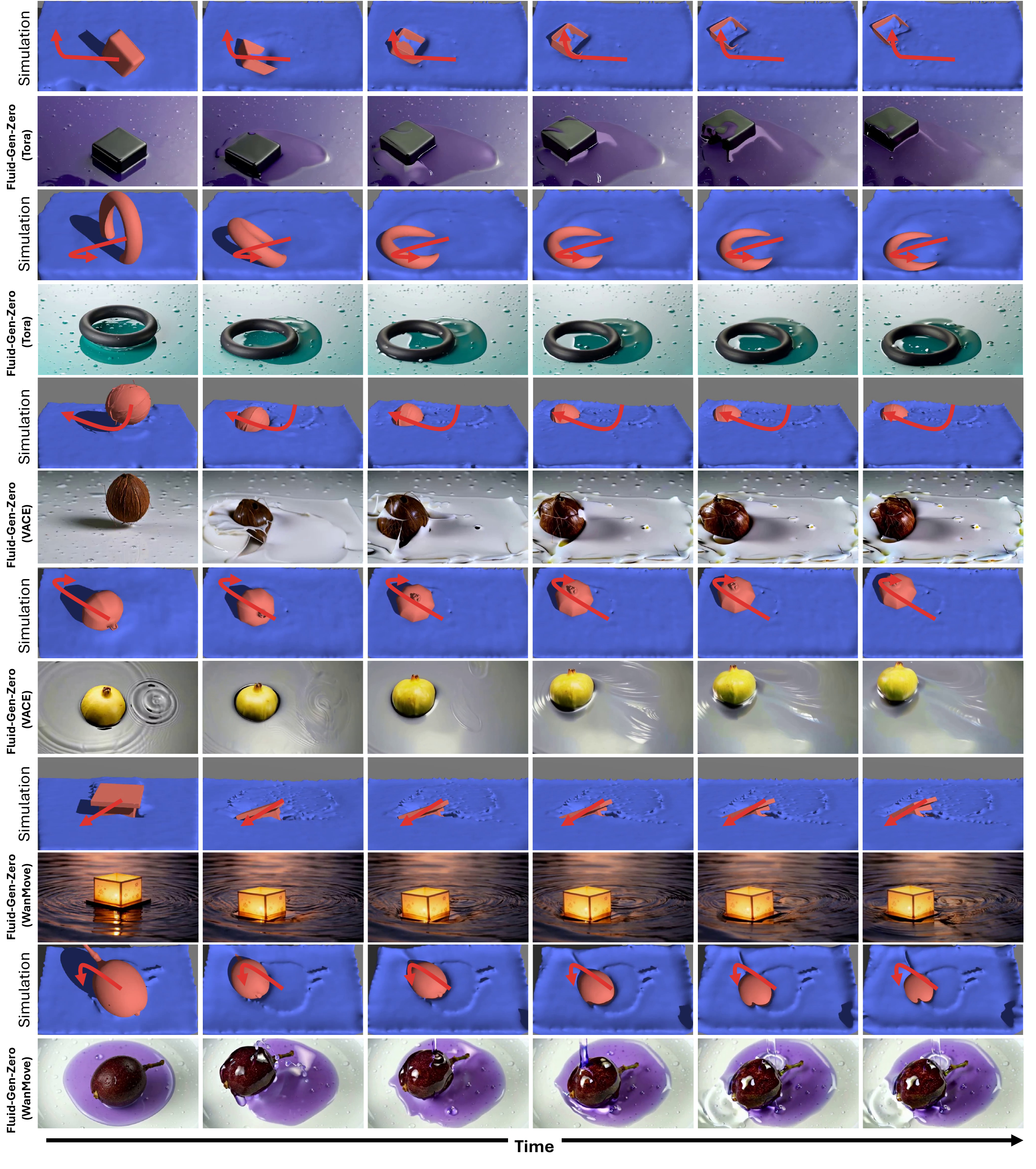}
    \caption{\textbf{Additional Fluid-Gen-Zero results.}}
    \label{fig:vis4}
\end{figure*}

\begin{figure*}[t]
\centering
\includegraphics[width=\linewidth]{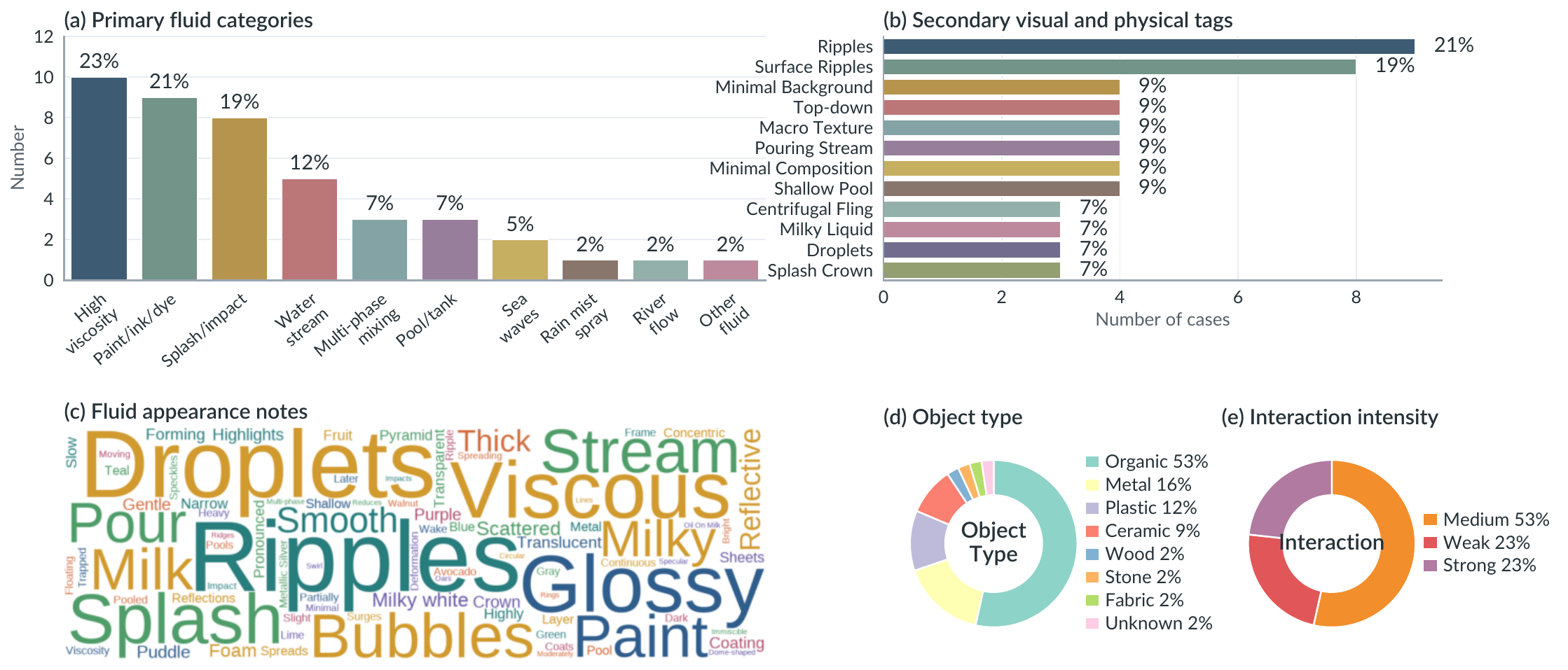}
\caption{\textbf{Benchmark statistics.}
Detailed distributions are computed from the diversity annotations of our data.}
\label{fig:benchmark_internal_statistics}
\end{figure*}

\begin{table*}[t]
\centering
\caption{\textbf{Comparison with related benchmarks and simulation-based data.}
Motion-control benchmarks provide visual trajectory annotations, while simulation-based systems expose richer physical signals.
Our benchmark is designed for fluid-object interaction evaluation with mask, 3D particle, and optical-flow references.}
\label{tab:benchmark_statistics}
\resizebox{\textwidth}{!}{
\begin{tabular}{lcccccc}
\toprule
\multirow{2}{*}{\textbf{Benchmark / Data}} &
\multicolumn{3}{c}{\textbf{Statistics}} &
\multicolumn{3}{c}{\textbf{Annotations}} \\
\cmidrule(lr){2-4}\cmidrule(lr){5-7}
&
\textbf{Samples} &
\textbf{Frames} &
\textbf{Categorization} &
\textbf{Mask} &
\textbf{3D Points} &
\textbf{Optical Flow} \\
\midrule
MagicBench~\cite{li2025magicmotion} &
600 &
49 &
\xmark &
\cmark &
\xmark &
\xmark \\
MoveBench~\cite{chu2025wanmove} &
1,018 &
81 &
\cmark &
\cmark &
\xmark &
\xmark \\
RealWonder~\cite{liu2026realwonder} &
30 &
--&
\xmark&
\xmark &
\cmark &
\cmark \\
PerpetualWonder~\cite{zhan2026perpetualwonder} &
10 &
--&
\xmark&
\xmark &
\cmark &
\cmark \\
\textbf{Ours} &
43 &
\makecell{\textbf{401 (simulation rollout)}} &
\cmark &
\cmark &
\cmark &
\cmark \\
\bottomrule
\end{tabular}}
\end{table*}

\section{Implementation Details}
\label{app:implementation_details}

\noindent\textbf{Perception and reconstruction.} Open-vocabulary segmentation uses Grounding DINO Base~\cite{liu2023grounding} followed by SAM 2.1 Hiera Large~\cite{ravi2024sam2}. Scene geometry comes from a monocular point map~\cite{moge2}, and the object mesh from an image-conditioned 3D model~\cite{sam3dobjects} given that same point map. Absolute scale is set by the object size estimated from the input image by the VLM.

\noindent\textbf{Warped reference.} SAM2 refines the first-frame object mask from the simulated mask centroid, and a FLUX-based inpainter~\cite{alimama2024fluxcontrolnetinpainting} synthesizes the background behind the removed object, using a neutral background prompt, $24$ diffusion steps, and a fixed seed.

\noindent \textbf{Simulation.} We use Genesis with the object represented as either a rigid body or an elastic MPM material, selected per scene by the VLM from the image and prompt, and the fluid represented as a liquid MPM material. A rigid object keeps its reconstructed mesh and couples to the liquid through the mesh's signed distance field, whereas an elastic object is filled with MPM particles. The time step is $\Delta t=10^{-3}$, with $30$ substeps and gravity set to $9.8\,\mathrm{m/s^2}$. Object elasticity, fluid density and viscosity, and force direction and magnitude come from the deterministic rule-based parser.

\noindent \textbf{VLM.} \texttt{gpt-5.4-mini-2026-03-17} for all three calls (intent parsing, risk annotation, prompt refinement), with frames sampled every five frames and no fine-tuning. If the API is disabled, the system falls back to the rule-based parser and the metric-only planner.

\noindent \textbf{GPT judge.} The GPT-based evaluation of physical realism (PhysR), photorealism (PhotoR), and semantic consistency (Sem) uses \texttt{gpt-5.5}.

\noindent \textbf{Latent-space guidance.} A single setting is shared by all backbones and all experiments: $k_0=8$, $k_{\mathrm{obj}}=11$, $k_{\mathrm{fl}}=10$, $\beta^f_{\max}=0.8$, and a Gaussian-smoothed fluid mask. Object-region wrapping therefore persists one step longer than fluid-region wrapping. None of these were tuned per scene or per backbone.

\noindent \textbf{Backbone routing.} Only the final preprocessing step differs. VACE receives the optical-flow condition and anchor frame in its native source-video/source-mask format; Tora and WanMove receive sparse motion trajectories derived from the same simulation rollout. All three backbones receive the warped reference video and masks as latent-control inputs. Clip lengths respect each backbone's valid frame counts, and no generator parameters are updated.

\section{Benchmark Size and Evaluation Scope}
\label{app:benchmark_scope}

Our benchmark contains 43 curated fluid-object interaction samples, all of which are used for the experiments in the main paper.
Each sample pairs an initial image and caption with a simulation rollout containing 3D particle trajectories, object and fluid masks, projected simulation optical flow, a simulation preview, and an appearance-preserving warped reference video.

\noindent\textbf{Why existing benchmarks do not cover it.}
\begin{itemize}[leftmargin=1.4em, itemsep=2pt, topsep=3pt]
    \item \textbf{Motion-control benchmarks} evaluate whether a generated video follows visual motion annotations. MagicBench provides object masks and bounding boxes~\cite{li2025magicmotion}; MoveBench provides category-balanced videos with first-frame masks and object point tracks~\cite{chu2025wanmove}. Both annotate image or video space, and so describe neither fluid state, material-dependent contact, nor coupled object-fluid evolution.
    \item \textbf{Simulation-based benchmarks} are closer in spirit but target different goals. RealWonder~\cite{liu2026realwonder} evaluates 30 action-conditioned samples for real-time generation from a single image, and PerpetualWonder~\cite{zhan2026perpetualwonder} studies closed-loop long-horizon 4D generation on 10 curated scenes.
\end{itemize}

A conventional entry is a prompt-video pair; each of ours additionally requires image understanding, 3D scene recovery, material assignment, object-fluid initialization, a 401-frame simulation rollout, projected masks and simulation optical flow, generation across three backbones, and comparison against that rollout. This process provides the dense 3D references needed to measure fluid motion.

The benchmark targets challenging interaction regimes, including contact, sliding, rotation, impact, submersion, splashing, wake formation, occlusion, viscous deformation, and long-horizon drift. These regimes stress the physical consistency of pretrained generators. Every sample runs through the same reconstruction, rollout, planning, and guidance procedure with no per-scene retuning, so the benchmark measures whether the same configuration generalizes across diverse challenging cases.

\cref{tab:benchmark_statistics} compares our benchmark with related benchmarks and simulation-based systems, and \cref{fig:benchmark_internal_statistics} visualizes the internal diversity of our annotations.

\begin{table*}[t]
\centering
\small
\setlength{\tabcolsep}{3.5pt}
\renewcommand{\arraystretch}{0.95}
\caption{\textbf{Effectiveness of complexity-aware planning on Tora.}
We compare backbone baselines with \textbf{uniform clip segmentation} against variants with generation-time planning.}
\label{tab:planning_effectiveness}
\begin{tabular}{lcccccccc}
\toprule
Method
& Size
& \makecell{Obj. Traj.\\Err. $\downarrow$}
& \makecell{Fluid\\fEPE $\downarrow$}
& \makecell{Subj.\\Cons. $\uparrow$}
& \makecell{Temp.\\Flick. $\uparrow$}
& \makecell{Aesth.\\Qual. $\uparrow$}
& \makecell{Img.\\Qual. $\uparrow$}
& \makecell{Physical\\Commonsense $\uparrow$} \\
\midrule

Tora
& 5B
& 133.20
& 3.950
& 0.804
& 0.980
& 0.487
& \textbf{0.656}
&  4.512 \\

Tora w/ Planning
& 5B
& \textbf{128.12}
& \textbf{3.221}
& \textbf{0.816}
& \textbf{0.982}
& 0.487
& 0.653
& \textbf{4.605} \\

\bottomrule
\end{tabular}
\end{table*}
\section{Additional Results}
\label{app:additional_results}

\subsection{Effectiveness of Complexity-Aware Clip Planning}

We first isolate the contribution of generation-time planning by comparing complexity-aware clip planning against uniform clip segmentation while disabling simulation-guided latent-space guidance.
This experiment changes only how the long simulation rollout is decomposed into generation clips and how clip-level prompts are assigned.

\begin{figure}[t]
    \centering
    \includegraphics[width=\linewidth]{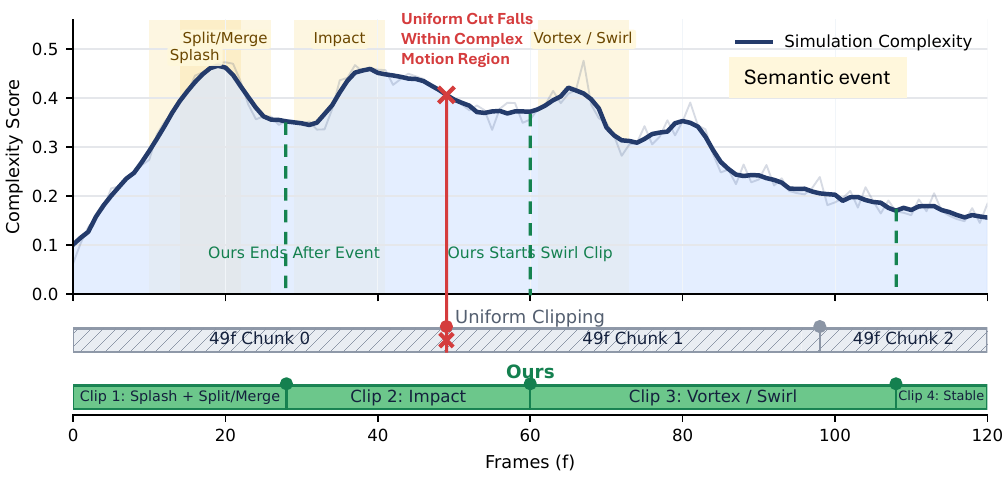}
    \caption{\textbf{Planned \emph{vs.}\ uniform clip boundaries}. A uniform 49-frame cut falls inside a complex motion region, whereas the planned boundaries end the first clip after the splash and start a new clip before the swirl.}
    \label{fig:planning_insight}
\end{figure}

As shown in \cref{tab:planning_effectiveness}, planning alone improves long-horizon simulation alignment and temporal consistency on Tora.
Fluid fEPE decreases by \textbf{18.5\%}, while object trajectory error also decreases from $133.20$ to $128.12$.
Subject consistency, temporal flickering, and physical commonsense improve as well, indicating that avoiding cuts through complex physical events helps preserve both motion continuity and event-level coherence.
\cref{fig:planning_insight} illustrates the difference between planned and uniform clip boundaries on a representative rollout.

\subsection{Effectiveness of Simulation-Guided Latent-Space Guidance}
\begin{table*}[h]
\centering
\vspace{-0.5em}
\small
\setlength{\tabcolsep}{3.0pt}
\renewcommand{\arraystretch}{0.95}
\caption{\textbf{Effectiveness of simulation-guided latent-space guidance on Tora.}}
\label{tab:latent_control_effectiveness}
\begin{tabular}{lcccccccc}
\toprule
Method
& Size
& \makecell{Obj. Traj.\\Err. $\downarrow$}
& \makecell{Fluid\\fEPE  $\downarrow$}
& \makecell{Subj.\\Cons. $\uparrow$}
& \makecell{Temp.\\Flick. $\uparrow$}
& \makecell{Aesth.\\Qual. $\uparrow$}
& \makecell{Img.\\Qual. $\uparrow$}
& \makecell{Physical\\Commonsense $\uparrow$} \\
\midrule

Tora
& 5B
& 133.20
& 3.950
& 0.804
& 0.980
& \textbf{0.487}
& \textbf{0.656}
&  4.512 \\

Tora w/ Latent-Space Guidance
& 5B
& \textbf{24.79}
& \textbf{0.812}
& \textbf{0.912}
& \textbf{0.995}
& 0.478
& 0.653
& \textbf{4.628} \\

\bottomrule
\end{tabular}
\end{table*}

We next isolate simulation-guided latent-space guidance by disabling complexity-aware clip planning.
Unlike generation-time planning, which determines how a long simulation rollout is temporally decomposed, latent-space guidance directly constrains the denoising trajectory within each clip.

As shown in \cref{tab:latent_control_effectiveness}, latent-space guidance substantially improves simulation alignment on Tora, reducing object trajectory error by \textbf{81.4\%} and fluid fEPE by \textbf{79.4\%}.
The improvement comes from anchoring the latent trajectory to the noised reference latent during early denoising instead of relying solely on the native backbone condition.
Subject consistency and temporal consistency improve substantially, while aesthetic and image quality change only slightly.
The asymmetric treatment of object and fluid regions balances physical guidance and generative freedom: strong object anchoring preserves position, silhouette, and contact structure, whereas softer fluid-region wrapping allows more flexibility for fluid appearance synthesis.

\subsection{Ablation Studies}
We also ablate two region-aware latent wrapping choices on the Tora backbone.
\textbf{w/o Region-Aware Wrapping} removes the spatial distinction between object and fluid regions and applies one wrapping schedule to both.
\textbf{w/o Fluid Wrapping} disables fluid-region wrapping while keeping the native backbone condition and object-region wrapping unchanged.

\begin{table*}[h]
\centering
\small
\setlength{\tabcolsep}{3pt}
\caption{\textbf{Ablation study of latent-space guidance on Tora.}
We evaluate key choices in region-aware latent wrapping.
}
\label{tab:ablation_latent_control}
\begin{tabular}{lccccccc}
\toprule
Setting
& \makecell{Obj. Traj.\\Err. $\downarrow$}
& \makecell{Fluid\\fEPE  $\downarrow$}
& \makecell{Subj.\\Cons. $\uparrow$}
& \makecell{Temp.\\Flick. $\uparrow$}
& \makecell{Aesth.\\Qual. $\uparrow$}
& \makecell{Img.\\Qual. $\uparrow$}
& \makecell{Physical\\Commonsense $\uparrow$} \\
\midrule
Full
& \textbf{24.61}
& \textbf{0.843}
& \textbf{0.906}
& 0.995
& \textbf{0.506}
& 0.656
& 4.628 \\

w/o Region-Aware Wrapping
& 30.40 & 1.026 & 0.882 & 0.985 & 0.476 & \textbf{0.657 }& 4.628 \\
w/o Fluid Wrapping
& 28.06 & 0.937 & 0.897 & \textbf{0.998} & 0.474 & 0.652 & 4.605 \\
\bottomrule
\end{tabular}
\end{table*}
\cref{tab:ablation_latent_control} confirms the importance of region-aware latent wrapping.
The full design achieves the best object trajectory error, fluid fEPE, subject consistency, and aesthetic quality, showing that treating object and fluid regions differently improves simulation alignment without substantially sacrificing appearance.
Removing region-aware wrapping degrades both object and fluid alignment, increasing object trajectory error from $24.61$ to $30.40$ and fluid fEPE from $0.843$ to $1.026$.
Disabling fluid wrapping also weakens simulation alignment, although it yields the highest temporal flickering score ($0.998$), suggesting that relaxing fluid constraints gives the pretrained generator more freedom to produce temporally smooth appearance.
Overall, these results show that stronger region-specific guidance improves simulation alignment, while relaxing the constraints can benefit individual perceptual metrics.

\begin{figure}[h]
    \centering
    \includegraphics[width=\linewidth,height=0.78\textheight,keepaspectratio]{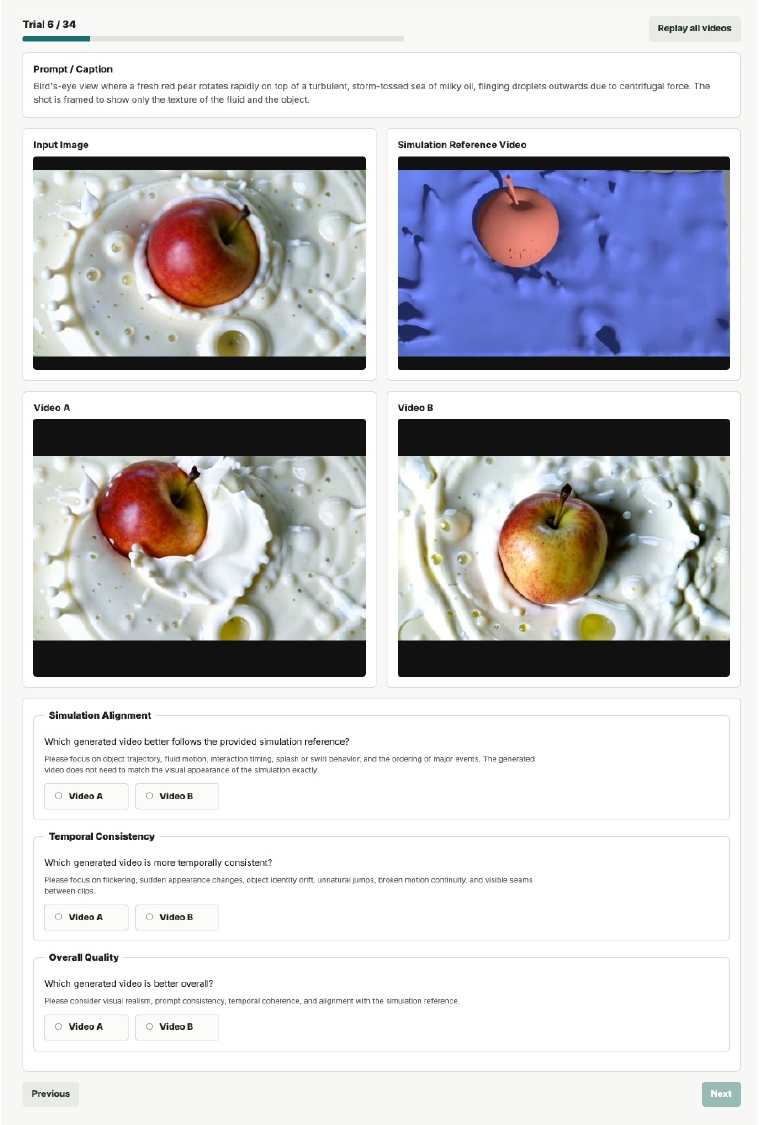}
    \caption{\textbf{User study screen.}
    Participants are shown the prompt, input image, simulation preview, and two anonymized generated videos.
    They answer three pairwise preference questions for simulation alignment, temporal consistency, and overall quality.}
    \label{fig:user_study_screen}
\end{figure}

\subsection{Qualitative Results}
\label{app:qualitative_results}

\begin{figure*}[t]
    \centering
    \includegraphics[width=\linewidth]{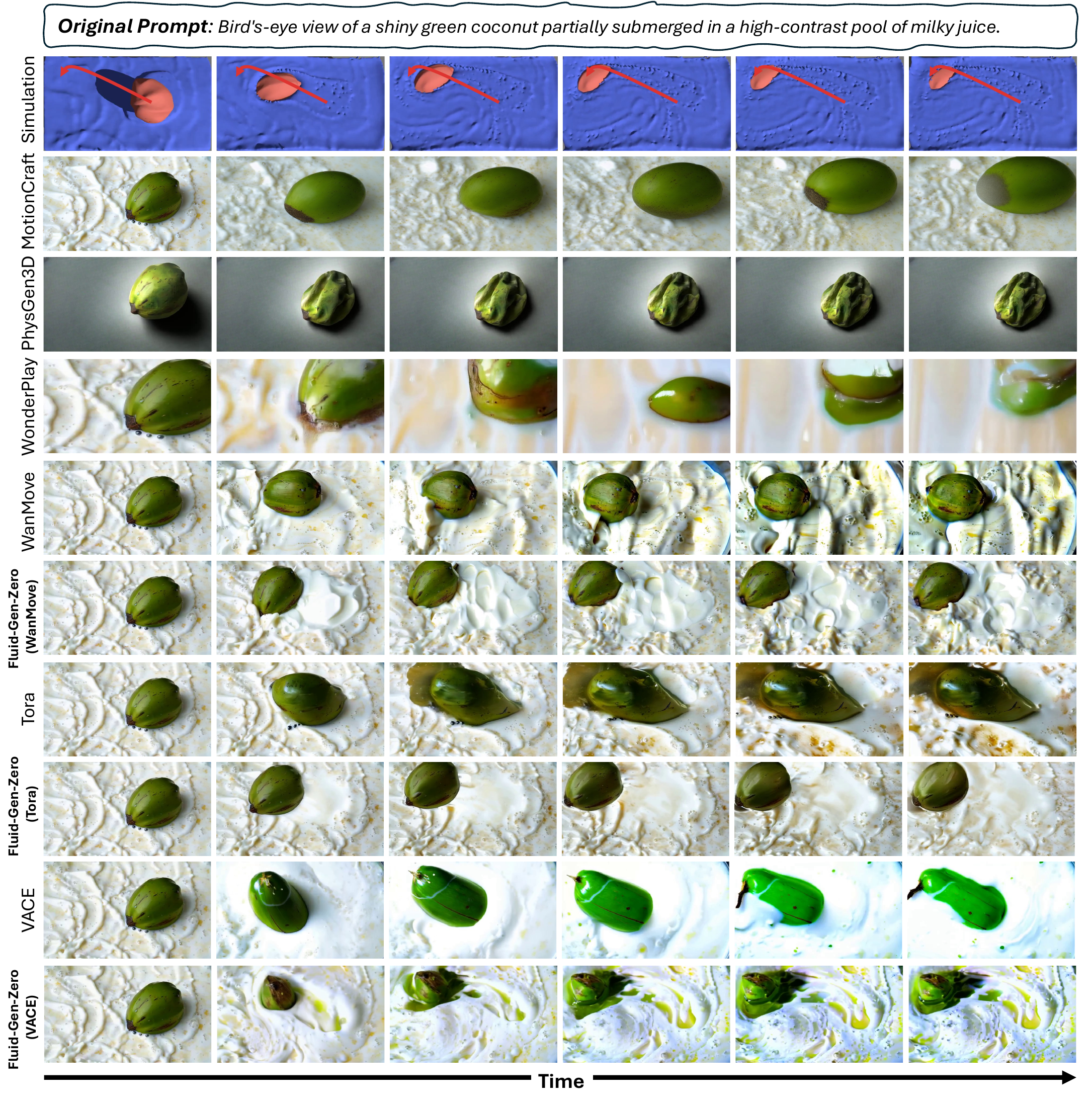}
    \caption{\textbf{Additional qualitative comparison results.}}
    \label{fig:vis3}
\end{figure*}

We provide additional qualitative comparisons across different scenes. In each sequence, the simulator supplies the coarse physical event order, object trajectory, and fluid spatial extent, while the video generator supplies photorealistic appearance. The comparisons evaluate whether the generated videos preserve these physically important motion cues while avoiding the appearance artifacts and temporal drift of the baseline generators.
\cref{fig:vis3} shows different methods' results, and \cref{fig:vis4} collects additional object-fluid cases with different materials, camera viewpoints, and interaction types.

\paragraph{Failure Case.}
\label{app:failure_case}

\begin{figure*}[t]
    \centering
    \includegraphics[width=\linewidth]{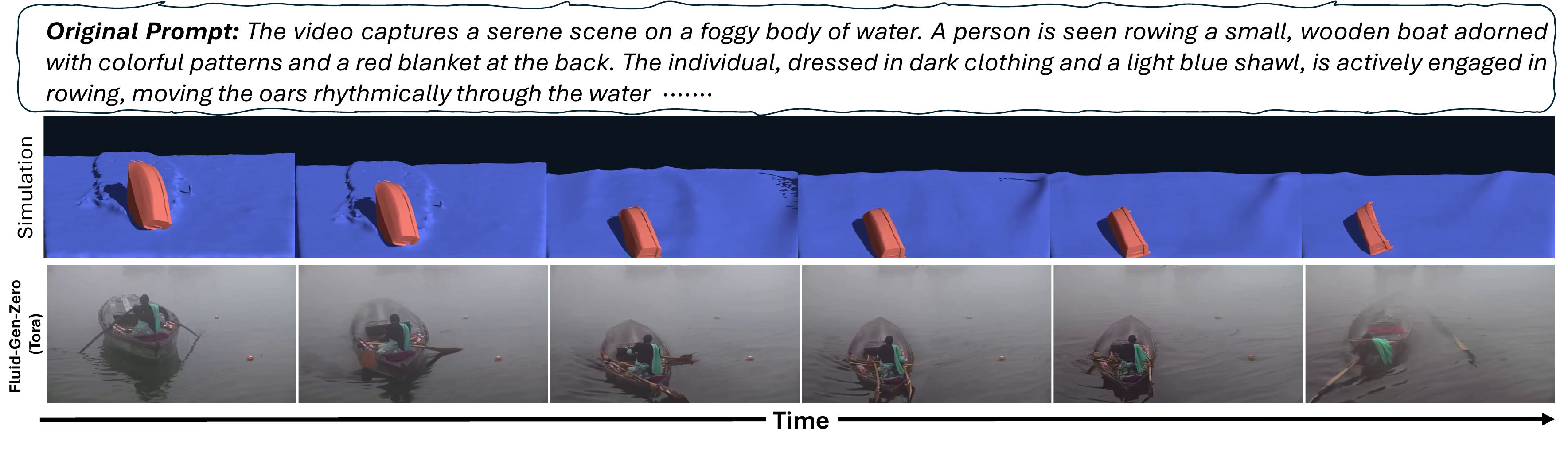}

    \caption{\textbf{Failure case.}
    When reconstruction or simulation is misaligned with the intended scene, Fluid-Gen-Zero may preserve an incorrect coarse spatial layout and produce contact or boundary artifacts.}
    \label{fig:failure_case}
\end{figure*}

\cref{fig:failure_case} shows a representative failure case of Fluid-Gen-Zero.
The most common failures occur when the reconstructed simulator state is physically plausible but not sufficiently aligned with the intended image-prompt scene.
For example, segmentation leakage, inaccurate depth near transparent fluid, imperfect object reconstruction, or incorrect pose and scale alignment can produce a simulation rollout whose object-fluid contact timing differs from the visual prompt.
In such cases, Fluid-Gen-Zero may still avoid cutting through protected event intervals and inject the noised reference latent, but the generated video can inherit an incorrect coarse spatial layout, show delayed or excessive fluid response, or produce boundary artifacts around contact regions.

\section{User Study Details}
\label{app:user_study_details}
\paragraph{Protocol.}
We conduct a two-alternative forced-choice (2AFC) user study on 8 samples.
For each sample, participants are shown the text prompt, input image, simulation preview, and two anonymized generated videos.
Method names are hidden and the order of Video A/B is randomized.
Each participant completes 34 pairwise trials: six samples are used for the three same-backbone comparisons and eight samples for the two comparisons against simulation-based methods.
We collect responses from 13 participants, resulting in 442 pairwise trials and 1326 criterion-level judgments. \cref{fig:user_study_screen} shows the screen used in the study.

\paragraph{Evaluation Instructions.}
Participants answer three preference questions for each pair.
For \emph{simulation alignment}, they are instructed to focus on object trajectory, fluid motion, interaction timing, splash or swirl behavior, and the order of major events. The generated video does not need to match the visual appearance of the simulation preview exactly.
For \emph{temporal consistency}, they are instructed to focus on flickering, sudden appearance changes, object identity drift, unnatural jumps, broken motion continuity, and visible stitching artifacts.
For \emph{overall quality}, they are instructed to consider visual realism, prompt adherence, temporal consistency, and alignment with the simulation preview.

\end{document}